%% file: sn-article.tex
\documentclass[sn-mathphys]{sn-jnl}

\jyear{2022}%

\theoremstyle{thmstyleone}%
\newtheorem{theorem}{Theorem}
\theoremstyle{thmstyletwo}%
\newtheorem{example}{Example}%

\theoremstyle{thmstylethree}%
\newtheorem{definition}{Definition}%

\renewcommand{\R}{\mathbb{R}} 
\renewcommand{\Z}{\mathbb{Z}} 
\newcommand{\N}{\mathbb{N}} 
\newcommand{\F}{\mathcal{F}}

\usepackage[utf8]{inputenc}
\usepackage[T1]{fontenc}
\usepackage[english]{babel}
\usepackage{hyperref}
\usepackage{setspace}

\usepackage{latexsym}
\usepackage{amsmath,amssymb,amsthm}
\usepackage[export]{adjustbox}
\usepackage{array}
\usepackage{tabularx}
\usepackage{mwe}
\usepackage[vcentermath]{youngtab}

\usepackage{calc}
\usepackage{import}
\usepackage{graphicx}
\usepackage{xcolor}
\definecolor{darkblue}{rgb}{0,0,0.5} 
\usepackage{transparent}
\usepackage{caption}
\usepackage[nottoc,notlot,notlof]{tocbibind}
\usepackage{ulem}

\begin{document}

\title[Equivariant and Steerable Neural Networks]{Equivariant and Steerable Neural Networks \\ {\large --  A review with special emphasis on the symmetric group --}}


\author*[1,2]{\fnm{Patrick} \sur{Krüger}}\email{patrick.krueger@uni-wuppertal.de}

\author[1,2]{\fnm{Hanno} \sur{Gottschalk}}\email{hanno.gottschalk@uni-wuppertal.de}

\affil*[1]{\orgdiv{Department of Mathematics and Science}, \orgname{University of Wuppertal}, \orgaddress{\street{Gaussstr. 20}, \city{Wuppertal}, \postcode{42110}, \country{Germany}}}

\affil[2]{\orgdiv{IZMD}, \orgname{University of Wuppertal}, \orgaddress{\street{Liese Meitner Str. 27-31}, \city{Wuppertal}, \postcode{42110}, \country{Germany}}}



\abstract{Convolutional neural networks revolutionized computer vision and natrual language processing. Their efficiency, as compared to fully connected neural networks, has its origin in the architecture, where convolutions  reflect the translation invariance in space and time in pattern or speech recognition tasks. Recently, Cohen and Welling have put this in the broader perspective of invariance under symmetry groups, which leads to the concept of group equivaiant neural networks and more generally steerable neural networks. In this article, we review the architecture of such networks including  equivariant  layers and filter banks, activation with capsules and group pooling. We apply this formalism to the symmetric group, for which we work out a number of details on representations and capsules that are not found in the literature.    }




\maketitle

\section{Introduction}
Neural networks are machine learning algorithms that are used in a wide variety of applications, for example in image recognition and language processing. There are many applications, among them automated communication in the service sector, perception for self driving cars and the interpretation of medical images in healthcare. In many machine learning problems, it is desirable to make the predictions of a network \textit{invariant} to certain transformations of the input. This means that, for instance in image classification, we want an image that was transformed by a rotation by some angle to still be classified with the same label: A picture of a cat rotated by 90 degrees is still a picture of a cat.\\
\ \\
An important step towards learning these \textit{invariant representations} was made by the introduction of convolutional neural networks by LeCun et al. for image classification of handwritten digits \cite{lecun1989backpropagation,lecun1998gradient}. These networks rely on the mathematical \textit{convolution operation} to achieve higher efficiency by \textit{parameter sharing}, as well as \textit{equivariance to translations}. Equivariance means that a shift in the input data in some direction will be carried through all but the ultimate fully connected layers of the network and result in a similar shift in further deep layers, which can then be used to achieve translation invariant representations.\\
\ \\
While convolutional neural networks thus yield some of the desired invariance, their performance still decreases when the data is transformed by other symmetries, e.g. rotations or reflections. While this can be avoided by data augmentation, another compelling approach is to extend the translation equivariance of convolutional networks to a wider class of transformations. This is realized by group equivariant neural networks, or, in short, $G$-CNNs, which were introduced by Cohen \& Welling in \cite{cohen2016group}. $G$-CNNs use group theory to generalize the convolution operation of conventional CNNs to \textit{group convolution}, which then yields equivariance not only to translations, but to a wider group of transformations $G$ of e.g. compositions of translations, reflections and rotations. These $G$-CNNs are again generalized by steerable CNNs (Cohen \& Welling, \cite{cohen2016steerable}) which, instead of modifying the convolution operation, instantiate \textit{equivariant filter banks} and \textit{steerable feature spaces} by making use of the theory of \textit{group representations}. They thereby achieve a more general and versatile concept of group equivariance.\\ 
\noindent
The aim of this article is to give an overview of the theory of equivariant networks. For this, some understanding of the mathematical theory of groups and group representations is needed, which we provide in chapter \ref{sectionmathprelims}.

Chapter \ref{sectiontheoryeqnets} will then be concerned with the theory of equivariant networks. After a short introduction on conventional, fully connected networks in section \ref{nns}, basic knowledge about convolutional neural networks is provided in section \ref{cnnbasics} by explaining the core concepts of feature maps, filters, the convolution operation and the translation equivariance resulting from it, as well as parameter sharing. In section \ref{sectiongcnn}, we will explain how a modification of the conventional convolution operation leads to $G$-CNNs and how these thereby achieve equivariance to groups of transformations. The more general theory of steerable CNNs will then be presented in section \ref{sectionsteercnns}. We will furthermore recapitulate how steerable CNNs are in fact a direct generalization of $G$-CNNs by explaining the equivalence of the latter to a special case of the former. The second aim of this article is to give some contributions by applications of $G$-CNNs and steerable CNNs for the symmetric group. This will be the focus of chapter \ref{sectionsnapplications}.

\section{Related Work}
Steerable CNNs have besides \cite{cohen2016steerable} been investigated by several other publications. Weiler et al. (\cite{weiler20183d}) discuss 3D-steerable CNNs, i.e. networks which process volumetric data with domain $\R^3$. In \cite{cohen2018intertwiners}, Cohen et al. give a more abstract approach to intertwiners in steerable CNNs and it is shown that layers of an equivariant network in fact \textit{need} to transform according to an induced group representation. A more general theory of steerable CNNs on homogeneos spaces is discussed by the same authors in \cite{cohen2018general}, showing that linear maps between feature spaces are in one-to-one correspondence to convolutions with equivariant kernels. $E2$-equivariant steerable CNNs that are equivariant to isometries of the plane $R^2$ in the form of continuous rotations, reflections and translations are discussed in \cite{weiler2019general}. Furthermore, gauge equivariant networks, which enable equivariance not just to global symmetries, but also to local transformations, are discussed by Cohen et al. in \cite{cohen2019gauge}.\\
\ \\
$G$-CNNs, i.e. networks that implicitly or explicitly rely on the regular representation of their respective group have also been studied several times. In \cite{gens2014deep}, Gens \& Domingos present symnets, which are a framework for networks with feature maps over arbitrary symmetry groups. Kanazawa et al. present in \cite{kanazawa2014locally} a way to learn scale invariant representations while keeping parameter cost low. Dielemann et al. discuss exploiting rotation symmetry to predict galaxy morphology in \cite{dieleman2015rotation} and expand their work to cyclic symmetries in \cite{dieleman2016exploiting}.
Another network architecture that relies on regular representations is given by scattering networks, which were defined by Mallat et al. in \cite{mallat2012group} for compact groups, as well as for the Euclidean group in \cite{oyallon2015deep} and \cite{sifre2013rotation}. \\
\ \\
Another important part of research concerning neural networks in general and equivariant networks specifically is the development of \textit{universal approximation theorems}. Many of these have been proved for different kinds of networks in varying levels of generality, with some of the first being \cite{hornik1989multilayer} and \cite{cybenko1989approximation}. Generally speaking, all approximation theorems state that neural networks can, under certain conditions, approximate any function from a predefined function space with arbitrary precision. In particular, it was proved by Leshno et al. in \cite{leshno1993multilayer} that multilayer feedforward networks can approximate any continuous function as long as non-polynomial activation functions are used. \\
Petersen and Voigtlaender showed in \cite{petersen2020equivalence} that fully connected feedforward networks can under minimal conditions be translated into convolutional (i.e. translation equivariant) networks, thereby enabling the application of most approximation theorems for the former to the latter. For group equivariant maps, Kumagai \& Sannai (\cite{kumagai2020universal}) also employ a conversion theorem from feedforward networks to CNNs to establish a method for obtaining universal approximation theorems for group equivariant convolutional networks.

\section{Mathematical Preliminaries}\label{sectionmathprelims}
In this section, some algebraic concepts that will be used and referenced throughout this thesis will be explained.\\
Secondly, we briefly recall the representation theory of finite groups.\\
Lastly, a short summary of the explicit representation theory of $S_n$, the symmetric group of $n$ letters, is presented. 

\subsection{Semidirect Products}
We define the core mathematical concepts needed for the theory of equivariant networks, starting with definitions of  semi-direct products. While this section is kept quite short, there exist many introductions to this topic, see g.g. \cite{bogopolski2008introduction} or \cite{hilton1908introduction}.\\

\begin{definition}\label{defgroupaction}\ \\
Let $(G,\circ)$ be a group and $X$ be some set. A \textit{(left) action} of $G$ on $X$  is a binary operation 
\begin{align}
\cdot: G\times X\to X,\,(g,x)\mapsto g\cdot x,
\end{align}
such that the following requirements are met:
\begin{itemize}
	\item [i)]$e\cdot x=x\;\forall x\in X$, with $e$ being the neutral element of $G$.
	\item[ii)] $(g\circ h)\cdot x=g\cdot (h\cdot x)\;\forall g,h\in G,x\in X$.
\end{itemize}
A group action is also often interpreted as a map 
\begin{align}
\phi_G:G\to 
\text{Aut}(X)=\{f:X\to X\mid f\text{ is bijective }\},\; g\mapsto\phi_g .
\end{align} 
Then, the two conditions above translate to the following:
\begin{itemize}
	\item [i)] $\phi_e=id$
	\item[ii)]$\phi_g\circ\phi_h=\phi_{gh}\;\forall g,h\in G$.
\end{itemize}	
\end{definition}

\begin{definition}\label{outersemidirectproduct}\ \\
	Let $(H,\circ_H)$ and $(N,\circ_N)$ be two groups and let furthermore $\phi_H:H\to \text{Aut}(N)$ be a group action of $H$ on $N$. Then, we can construct the \textit{outer} semidirect product $(N\rtimes H,\bullet)$ by setting the cartesian product $N\times H$ as the group's underlying set and defining the group operation as follows: 
	
	\begin{equation}
	\begin{gathered}
	\bullet:N\rtimes H\to N\rtimes H,\\
	(n_1,h_1)\bullet (n_2,h_2)=(n_1\circ_N\phi_H(h_1)n_2, h_1\circ_H h_2).
	\end{gathered}
	\end{equation}
	With this operation, $H\cong\{(e_N,h)\mid h\in H\}$ and $N\cong\{(n,e_H)\mid n\in N\}$ become subgroups of $(N\rtimes H)$, with the latter being a normal subgroup as
	\begin{align}
	(n,h)\bullet (n',e)=(n\circ_N\phi(h)n'\circ_N n^{-1},e)\bullet (n,h)\;\forall (n,h)\in N\rtimes H, (n',e)\in N.
	\end{align}
	The neutral element is given by $(e_N,e_H)$ and we have $(n,h)^{-1}=(\phi_H(h^{-1})n^{-1},h^{-1})$ for any $(n,h)\in H\rtimes N$.
	Furthermore, using above isomorphisms, any element $(n,h)\in N\rtimes H$ can be written as a unique product
	\begin{align}
	(n,h)=(n,e_H)\bullet (e_N,h).
	\end{align}
	As $N\cap H=(e_N,e_H)$ holds trivially, the outer semidirect product thus has all properties of the inner semidirect product with respective isomorphic subgroups. 
\end{definition}

\begin{example}\label{explgroupaction}\ \\
	I shall name two explicit examples for semidirect product groups and their respective actions here. These examples will be referenced in later parts of this article. 
	\begin{enumerate}

		\item [i)] The group $p4$ of compositions of rotations and translations of the square grid $\Z^2$ is the semidirect product of $\Z^2$ and $C_4$, the group of 90-degree-rotations around any origin. Elements of $p4$ can be parametrized by $3\times 3$ matrices depending on a rotational coordinate $r\in\{0,1,2,3\}$ and two translational coordinates $(u,v)\in\Z^2$:
		\begin{align}
		g(r,u,v)=\begin{pmatrix}
		\cos(\frac{r\pi}{2})&-\sin(\frac{r\pi}{2})&u\\
		\sin(\frac{r\pi}{2})& \cos(\frac{r\pi}{2})&v\\
		0&0&1
		\end{pmatrix}
		\end{align}
		$p4$ now acts on points $x=(u',v')\in\Z^2$ by matrix multiplication from the left, after adding a homogenous coordinate to $x$, i.e. $x=(u',v')\mapsto (u',v',1)$.
		\item[ii)] The group $p4m$ of compositions of rotations, mirror reflections and translations of $\Z^2$ is the semidirect product of $\Z^2$ and $D_4$, which is the group of 90-degree-rotations and reflections about any origin. Elements of this group can be parametrized similarly to $p4$, with the difference of adding a reflection coordinate $m$:
		\begin{align}
		g(m,r,u,v)=\begin{pmatrix}
		(-1^m)\cos(\frac{r\pi}{2})&-(-1^m)\sin(\frac{r\pi}{2})&u\\
		\sin(\frac{r\pi}{2})& \cos(\frac{r\pi}{2})&v\\
		0&0&1
		\end{pmatrix}
		\end{align}
		$p4m$ then acts on $\Z^2$ analogously to $p4$.
	\end{enumerate}
\end{example}
\noindent In section \ref{sectionsnapplications}, we will investigate steerable CNNs that rely on the semidirect product of the symmetric group $S_n$ and $\Z^n$. For this, some preliminary definitions shall also be given here. 
\newpage

\begin{definition}\label{defSn}
\phantom{l}
\begin{itemize}
    \item[i)]The symmetric group $S_n$ is the group of permutations of $n$ letters, i.e.
	\begin{align*}
	S_n=\{\sigma:\{1,..,n\}\to\{1,..,n\}\mid\,\sigma \text{ is bijective}\},
	\end{align*}
	with the group operation being defined by composition of elements, i.e. functions.
	\item[ii)] For $r\leq n$, a \textit{r-cycle} is an element of $\sigma \in S_n$, such that there exist\\ $x_1,..,x_r \in \{1,..,n\}$ with
		\begin{align*}
		\sigma(x_1)=x_2,\sigma(x_2)=x_3,..,\sigma(x_{r-1})=x_r, \sigma(x_r)=x_1,\\ \sigma(x_k)=x_k \;\forall\; x_k\notin \{x_1,..,x_r\}.
		\end{align*}
		An $r$-cycle as above is often denoted $(x_1\,x_2\,..\,x_r)$. We call this the \textit{cycle notation}, and we call $r$ the \textit{cycle length} of $\sigma$.
	\item[iii)] We obtain an action of $S_n$ on $\Z^n$ for any $n\in\N$ by letting $\sigma\in S_n$ permute the coordinates of $x=(x_1,..,x_n)\in\Z^n$:
\begin{align}\label{sigmatrans}
\sigma\circ (x_1,..,x_n)=(x_{\sigma(1)},..,x_{\sigma(n)})
\end{align}
\end{itemize}
	 
\end{definition}

\begin{example}\label{explsn}\ \\
    With above action, we can construct the outer semidirect product $\Z^n\rtimes S_n$ of products $t\sigma$ of translations $t$ and coordinate permutations $\sigma$. This group can now be parameterized and made to act on $\Z^n$ in analogy to example \ref{explgroupaction} by $n+1\times n+1$ matrices with the upper left $n\times n$ block being a permutation matrix representing $\sigma$ and a translation vector $t\in\Z^n$ in the first $n$ entries of the last column. Two examples for $n=2$ and $n=3$ shall be given here:
\begin{equation}
\begin{gathered}
g_2(((12),(3,3)))=
\begin{pmatrix}
0&1&3\\
1&0&3\\
0&0&1
\end{pmatrix},
g_3(((123),(1,2,2)))=
\begin{pmatrix}
0&0&1&1\\
1&0&0&2\\
0&1&0&2\\
0&0&0&1\\
\end{pmatrix}
\end{gathered}
\end{equation}
\end{example}

\subsection{An Introduction to Representation Theory of Finite Groups}\label{sectionreptheory}

\begin{definition}\label{defrepresentation}\ \\
	Let $G$ be a group.
	\begin{itemize}
		\item[i)] A \textit{representation} $(V,\rho)$ of $G$ consists of a vector space $V$, together with a group homomorphism  $\rho:G\to GL(V)$ from $G$ to the general linear group of $V$, i.e. the group of invertible linear maps from $V$ to $V$, such that:
		\begin{align}
		\rho(gh)=\rho(g)\rho(h)\, \forall g,h\in G.
		\end{align}
		The \textit{dimension} or \textit{degree} of the representation is defined as the dimension of its vector space $V$. 
		\item[ii)] A \textit{subrepresentation} of a representation $(V,\rho)$ is a subspace $W\subseteq V$ which is \textit{G-invariant}, meaning $\rho(g)w\in W$ for all $g\in G, w\in W$. The subrepresentation is then denoted by $(W,\rho\restriction_W)$, and we have
		
		\begin{align}
		\rho \restriction_W(g)=\rho(g)\restriction_W.
		\end{align}
		
		Any representation $(V,\rho)$ always has at least two subrepresentations: Itself and $\{0\}$.
		\item[iii)] If a representation $(V,\rho)$ with $V\neq\{0\}$ has no subrepresentations except itself and $\{0\}$, it is called \textit{irreducible}. Otherwise, it is called \textit{reducible}. Throughout this thesis, we will often refer to irreducible representations as \textit{irreps}.
		\item[iv)] For two representations $(V_1,\rho_1)$ and $(V_2,\rho_2)$ of $G$ of respective degrees $n_1$ and $n_2$ we can define the \textit{direct sum} of representations, $(V_1\oplus V_2,(\rho_1\oplus\rho_2))$, with\\ $(\rho_1\oplus\rho_2)(g)(v,w)=(\rho_1(v),\rho_2(w))$. This is again a representation of $G$ and has degree $n_1+n_2$.
	\end{itemize}
\end{definition}

\begin{example}[\textbf{The Quotient Representation}]\label{exquotrep}\ \\
	Given the quotient space (or quotient group) $G/H$ of a group $G$ and some subgroup $H\subseteq G$, we define the \textit{quotient representation} $(V_{\text{quot}},\rho_{\text{quot}})$ by associating a basis vector with every coset in $G/H$:
	\begin{align}
	V_{\text{quot}}=\langle\{e_{gH}\mid gH\in G/H\}\rangle
	\end{align}
	and define $\rho_{\text{quot}}$ using the action of $G$ on cosets: 
	\begin{align}
	\rho_{\text{quot}}(g')e_{gH}=e_{g'gH}\;\forall g'\in G,\;\forall e_{gH}\in V_{\text{quot}}.
	\end{align}
	As this representation just permutes the basis vectors, this time corresponding to cosets, this can also be realized by permutation matrices. If we choose $H={e}$, and thus receive $G/H=G$, this yields the so-called \textit{regular representation} which permutes basis vectors $e_g$ for each $g\in G$.
\end{example}

\begin{definition}\ \\
	Let $(V_1,\rho_1)$ and $(V_2,\rho_2)$ be two representations of some group $G$ and let $f:V_1\to V_2$ be a linear map. 
	\begin{itemize}
		\item [i)] $f$ is called an \textit{intertwiner} between $\rho_1$ and $\rho_2$, if 
		\begin{align*}
		f(\rho_1(g)(v))=\rho_2(g)(f(v)),\,\forall g\in G,\,\forall v\in V_1.
		\end{align*}
		\item[ii)] 	$\rho_1$ and $\rho_2$ are called \textit{equivalent} or \textit{isomorphic}, if there exists an intertwiner $f:V_1\to V_2$ between them, which also is a vector space isomorphism, i.e. an invertible linear map. We then often write $V_1\simeq V_2$ or $\rho_1\simeq \rho_2$.
		\item[iii)] The condition in i) is linear in $f$, thus any linear combination of intertwiners between representations $\rho_1$ and $\rho_2$ is again an intertwiner. We hence receive a vector space of intertwiners between $\rho_1$ and $\rho_2$, denoted $\text{Hom}_G(\rho_1,\rho_2)$.
	\end{itemize}	
\end{definition}

\begin{theorem}[\textbf{Schur's Lemma}]\label{schurslemma}\ \\
	Let $(V_1,\rho_1)$ and $(V_2,\rho_2)$ be irreducible representations of $G$.
	\begin{itemize}
		\item [i)] Iff $V_1$ and $V_2$ are not isomorphic, then dim $\text{Hom}_G(\rho_1,\rho_2)=0$, ie the only intertwiner between $\rho_1$ and $\rho_2$ is the zero map.
		\item[ii)] Iff $V_1$ and $V_2$ are isomorphic, then dim $\text{Hom}_G(\rho_1,\rho_2)=1$, and all maps intertwining $\rho_1$ and $\rho_2$ are scalar multiples of the identity map. 
	\end{itemize}
\end{theorem}

\begin{theorem}\label{thmirrepdecomp}
	\phantom{l}
	\begin{itemize}
		\item[i)] Any finite dimensional representation $(V,\rho)$ of a finite group can be decomposed into a direct sum of irreducible representations:
		\begin{align*}
		(V,\rho)\sim(\oplus V_i,\oplus \rho_i),
		\end{align*}
		where $(V_i,\rho_i)$ are irreps of $G$.
		\item[ii)] The irreps of $G$ are uniquely determined up to isomorphism, and there are finitely many of them.
		
	\end{itemize}
\end{theorem}
\begin{proof}
	See \cite{serre1977linear}, section 1.4, Theorem 2 for i) and section 2.5, Theorem 7 for ii).
\end{proof}

\begin{definition}\label{irreptype}\ \\
	Let $Irr(G)$ be the set of nonisomorphic irreducible representations of $G$. Then, any $(V_i,\rho_i)\in \text{Irr}(G)$ can occur in the decomposition of some finite dimensional representation $(V,\rho)$ any number of times. A list of integers $m_{\rho_i}(\rho)\geq 0$ corresponding to the multiplicity of each irrep $(V_i,\rho_i)$ in $(V,\rho)$ is called the \textit{type} of $\rho$. Representations are uniquely determined up to isomorphism by their types, meaning that if two representations have the same type, then they are isomorphic and that isomorphic representations always have the same type.\\	
\end{definition}

\section{The Theory of Equivariant Networks}\label{sectiontheoryeqnets}
In this section, the general theory behind convolutional neural networks is explained. We cover three types of convolutional networks which are subsequent generalizations of each other. In section \ref{nns}, We give a brief overview on how non-convolutional, fully connected networks work. Section \ref{cnnbasics} will then cover the basics on standard \textit{translation equivariant} convolutional networks. Key concepts, such as \textit{feature maps}, \textit{filters}, convolutional layers and the  convolution operation, as well as activation and pooling layers is discussed.
In section \ref{sectiongcnn}, we elaborate the generalization of convolutional networks to group equivariant networks (G-CNNs). All concepts from the previous section are  suitably generalized, such as e.g. $G$-feature maps and group convolution will be touched on, and it will be shown how the resulting networks have a ``more general'' form of equivariance.
Lastly, in section \ref{sectionsteercnns}, we introduce steerable convolutional networks, which use representation theory to yield a more efficient and versatile way to define group equivariant networks where the group acts on fibers of feature maps, also called filter banks.
It should be noted that while G-CNNs and steerable CNNs can be realized for many kinds of groups, the sections \ref{sectiongcnn} and \ref{sectionsteercnns} will focus on networks that use split groups, i.e. groups that are constructed as a semidirect product. Explicitly, we will consider networks on $p4$ and $p4m$, which were defined in example \ref{explgroupaction}, iv) and v). However, the concepts can easily be generalized to other split groups. A $G$-CNN that does rely on a non-split group will be covered in section \ref{sectionsnapplications}.
\subsection{Fully Connected Neural Networks}\label{nns} 
Deep neural networks describe a wide range of computing systems in the domain of machine learning that are meant to loosely resemble the human brain by using interconnected layers, indexed in the following by $l=1,...,L$, of  \textit{neurons}. Each neuron $N_i^l, i=1,...,K_l$ at each layer $l$ of the network can be thought of as a simple unit that holds a number, usually referred to as its \textit{activation}, and which is connected to all of neurons of the next layer. Therefore, this kind of network is often called \textit{fully connected}. Each of these connections is determined by a \textit{weight}  and a \textit{bias}. We denote by $\omega_{i,j}^l$ and $b_{i,j}^l$ the weight and bias that connect the $i$th neuron of layer $l$, $N_i^l$, to the $j$th neuron $N_j^{l+1}$ of layer $l+1$. In the so-called \textit{forward propagation step} of a network, starting at $l=1$, each neuron is updated in the following way:

\begin{align}
    N_i^l=\sigma_l\left(\sum_{j=1}^{K_{l-1}}\omega_{j,i}^{l-1}N_j^{l-1}+b_{j,i}^{l-1}\right),
\end{align}
\noindent
where $\sigma$ is some \textit{non-linear activation function}. These will be discussed in \ref{Nonlins_Pool}
This process is then repeated sequentially for all remaining $l=2,...,L$, until the final layer is reached. This \textit{output layer} consists of one neuron for each possible label of the classification problem at hand. The output neuron with the highest activation is returned as the networks predicted label for the given input.\\ 
\ \\
To train a deep neural network, a training dataset with inputs that are already labeled with their respective classifications is needed. During the training phase, after each forward pass, the (initially large) error of the network, i.e. the difference between the model's prediction and the actual labels of the training data is quantified by a \textit{loss function} which takes as inputs the weights and biases of the network. A \textit{backpropagation} algorithm then produces the gradients of this loss function with respect to its variables (i.e. the weights and biases), which are often called the \textit{parameters} of the networks. After obtaining the gradients, an optimization algorithm, the most used one being \textit{stochastic gradient descent} (\cite{bottou2010large}), is used to modify the networks parameters, thereby slightly optimizing the loss function. This training process of forward pass and backpropagation is now repeated until the error is sufficiently small. Then, the backpropagation step is no longer needed and the network can be used to classify unlabeled data. The reader interested in a more thorough introduction to feedforward networks is referred to chapter 6 of \cite{goodfellow2016deep}.

\subsection{Basics on Convolutional Neural Networks}\label{cnnbasics}
Convolutional Neural Networks (CNNs) are a powerful tool in machine learning, mainly used for pattern recognition and classification of certain input data. They can be used on a variety of data, for example sound signatures (one-dimensional data) and 2D and/or 3D Images. Just as fully connected networks, CNNs work in layers, in each of which a set of \textit{filters} representing certain features to look for in the input data is \textit{convolved} with stacks of so-called \textit{feature maps}. After application of certain operations known as \textit{nonlinearities} and \textit{pooling}, this yields a new set of feature maps, which is then convolved with a new set of filters in the next layer. A core feature of CNNs is the \textit{equivariance to translation} of their convolutional layers. This means that shifting the input data of a convolutional Layer will result in a similar shift that layer's output, allowing us to detect features independent of their location. In the next few paragraphs we will explain these terms in more detail. We use a Network operating on 2D black and white images, i.e. an input feature map with one channel as example.
\subsubsection{Feature Maps and Filters}
\textbf{Feature maps}\\
Feature maps are mathematical functions used to describe the input data, as well as the outputs of each convolutional layer. Depending on the type of the input, their domain can vary. In image recognition, the domain of a feature map usually is the two-dimensional pixel grid $\Z^2$.  In our example of a black and white image, an input-level feature map $f:\Z^2\to \R$ simply returns the grey-value at each pixel $x\in \Z^2$. In a coloured image, the function would instead be of the form $f:\Z^2\to \R^3$, returning a vector consisting of the color channel values at each pixel coordinate. Since images are bound in size, a feature map is usually said to just return zero everywhere outside of a certain subdomain of pixels. Since layers of convolutional networks contain more of one feature map most of the time, it is often also spoken of 'stacks' of feature maps $f^j:\Z^2\to\R^K, j=1,..,n$, whereby the index is often omitted for simplicity. \\

\noindent
\textbf{Filters}\\
Filters are used to extract certain characteristic patterns in our data by convolving them with feature maps, which will be described in the next paragraph. Mathematically, they are also described as functions. For convolution to work, it is important that these functions share the domain and image space of the feature maps that they are to be convolved with. In our example, a one-channel filter looking for diagonal lines (top left to bottom right) could look like this:
\begin{align}\label{explfilter}
\psi:\Z^2\to \R; \hspace{1cm} \psi(x)=\begin{cases}
1& x\in\{(-1,1),(0,0),(1,-1)\}\\
0& \text{elsewhere.}
\end{cases}
\end{align}
The support, i.e. the non-zero domain of filters is usually much smaller than that of the feature maps, with usual widths being $3\times 3$ or $5\times 5$ squares centered at the origin. While classical computer vision used handcrafted filters as in the above example, CNN based computer vision uses learnable filters $\psi=(\psi_{i,j}))_{i,j=-s,\ldots,s}$, $s=1,2$. The values $\psi_{i,j}$ are the learned parameters of the network. As will elaborated in \ref{ParamSharing}, this drastically reduces the parameter cost of CNNs in comparison to fully connected networks. \\

\subsubsection{The Convolution Operation}
The core building block of regular CNNs are the so called \textit{convolutional layers}. In each of these layers, a stack of feature maps $f:\Z^2\to \R^K$ is \textit{convolved} with a set of filters $\psi:\Z^2\to\R^K$, producing a new feature map which will then be used in the next layer. The convolution operation is mathematically defined as follows:
\begin{align}\label{convolution}
f\star\psi :\Z^2\to \R; \hspace{1cm}[f\star\psi](x)=\sum_{y\in\Z^2}\sum_{k=1}^{K}f_k(y)\psi_k(y-x)
\end{align}
While this may look complicated at first glance, it can be interpreted as simply sliding our small filter window over the domain of the image or feature map and summing up the element-wise products of the filter's values and the feature map's values lying ``under'' them at each output channel $k$ and each respective position $x$ of the filter. 
To demonstrate this further, we can interpret our example filter $\psi$ from (\ref{explfilter}), as well as a feature map $f$ as matrices. The feature map and filter, as well as the result of the convolution operation of said elements is illustrated in figure \ref{figure1}.\\
Here, as in all following figures throughout this thesis, black pixels represent one, grey pixels zero and white pixels represent minus one. \\
\begin{figure}[t!]
	\centering
	\def\svgscale{0.4}
	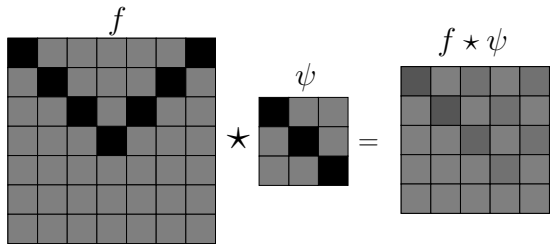
	\caption{A feature map $f$ representing the letter ``V'' is convolved with a filter detecting diagonal edges from top left to bottom right, yielding a new feature map $f\star\psi$. }
	\label{figure1}
\end{figure}

\noindent
Recall that, even though the domain of $f$ and $\psi$ is infinite, both functions just return 0 everywhere outside of the depicted areas. 

Convolving a feature map with a filter yields a new feature map $f\star\psi:\Z^2\to\R$, describing how well the feature described by the filter fits at different positions $x\in\Z^2$ in the image.\\
\ \\
As announced, we will from now on often omit the summation over channels in \eqref{convolution} for simplicity and pretend to have feature maps and filters with just one channel, as this will not harm any arguments made in the following paragraphs.\\

\subsubsection{Translation Equivariance}
An important property that makes CNNs such powerful tools, for example in image recognition, is their equivariance to translations of the input at each layer. Roughly speaking, this means that a CNN is able to detect features in an image (or other data) regardless of their specific position, without requiring additional parameters. To be more precise, we define a translation of a feature map mathematically:
\begin{align}
[T_tf](x)=f(t^{-1}x)=f(x-t);\; x,t\in\Z^2
\end{align}

\begin{figure}[h]
	\centering
	\def\svgscale{0.35}
	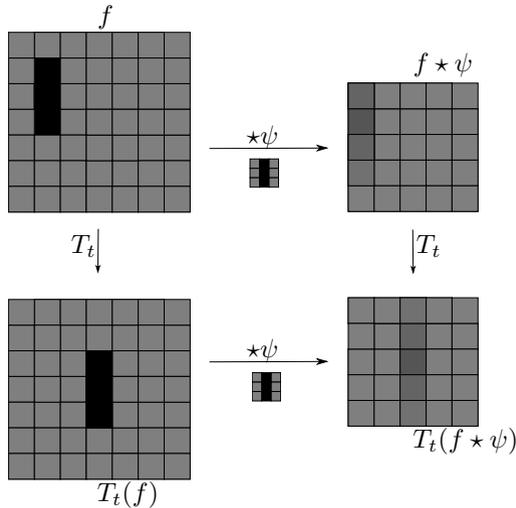
	\caption{A translation equivariant convolution. Here, the feature maps are transformed by $t=(-2,1)$ and the diagram commutes.}
	\label{figure2}
\end{figure}

\noindent
A feature map $f$ is hence transformed by a translation $t\in\Z^2$ by looking up the value of $f$ at the point $t^{-1}x=x-t$ and moving it to the position $x$, resulting in the $t$-transformed feature map $T_tf$. Visually, this translates to a shift of the patterns of a given input by the respective horizontal and vertical coordinates of $t$. For an illustration, see the left side of figure \ref{figure2}. The underlying concept of a group action of $\Z^2$ on itself is also used by the convolution operation itself: In (\ref{convolution}), filters are transformed by $T_x$ when calculating the output of the convolution at position $x$, representing a shift of said filter to that position.\\
\ \\
\noindent  
Equivariance to translations in CNNs now means that applying a translation $t$ to a feature map $f$, followed by a convolution with a filter $\psi$  yields the same result as convolving $f$ with $\psi$ first and then applying the translation:
\begin{equation}\label{transeq}
\begin{split}
[[T_tf]\star\psi](x)&=
\sum_{y\in\Z^2}f(t^{-1}y)\psi(x^{-1}y)\\
&=\sum_{y\in\Z^2}f(y-t)\psi(y-x)\\
&=\sum_{y\in\Z^2}f(y)\psi(y+t-x)\\
&=\sum_{y\in\Z^2}f(y)\psi(y-(x-t))\\
&=\sum_{y\in\Z^2}f(y)\psi((t^{-1}x)^{-1}y)\\
&=[T_t[f\star\psi]](x).\\
\end{split}
\end{equation}

\subsubsection{Parameter Sharing}\label{ParamSharing}
In comparison to conventional, fully connected neural networks (section \ref{nns}), convolutional neural networks increase the efficiency of their parameters by utilizing the convolution operation to achieve \textit{parameter sharing} across layers.\\
\ \\
In fully connected layers of a conventional neural network, each neuron is connected to the neurons of the next layer by an individual connection consisting of a weight and a bias, which together make up the parameters of the network. Each of these connections is modified individually to achieve the best possible classification result. In image recognition for instance, one can view each pixel of the input image as a neuron, and each of these is connected via individual parameters to the neurons of the subsequent layer. As layers are stacked on top of each other, one can imagine that the number of connections, and thus of parameters will grow quite large. \\
\ \\
For convolutional neural networks, consider each channel $f_l^{k'},\,k'=1,..,K^l$ of a feature map $f_l:\Z^2\to \R^{K^l}$ with $K^l$ output channels in a given layer $l$. Each of these channels is obtained by convolving a small filter patch $\psi^{k'}$ of size $s\times s\times K^{l-1}$ with the stack of feature maps $f_{l-1}^k$ of the previous layer. Here, $k=1,..,K^{l-1}$ denotes the individual channels and $K^{l-1}$ is the number of channels of the feature map of layer $l-1$. All values $f_l^{k'}(x)$ of an output channel $k'\in\{1,..,K_l\}$ for each $x\in\Z^2$ are thus obtained from the same filter, evaluated at different spatial coordinates, and hence \textit{share} the same $s^2K^{l-1}$ parameters. For the whole layer $l$, this then amounts to $s^2K^{l-1}K^l$ parameters. For comparison, as each pixel would need its own parameters, a fully connected layer processing the same data would need around $K^{l-1}K^l{Z^{l-1}}^2{Z^l}^2$ parameters with $Z^{l-1}$ and $Z^l$ being the spatial extend of the non-zero domains of the feature maps $f_{l-1}$ and $f_l$, which is substantially larger than $s^2$, with $s$ usually being 7 or smaller.  

\subsubsection{Nonlinearities and Pooling}\label{Nonlins_Pool}
\textbf{Nonlinearities}\\
The convolutional layers of a CNN are interspersed with non-linear layers. Such a layer operates by applying a non-linear function to the input is needed to enhance the expressive power of CNNs, since hardly any phenomenon that CNNs are meant to observe can be described  by linear functions. Various \textit{universal approximation theorems}, for instance in \cite{leshno1993multilayer} and \cite{petersen2020equivalence,yarotsky2022universal}, prove that (feedforward and convolutional) networks can approximate almost any function with arbitrary levels of precision, as long as non-polynomial nonlinearities are used. \\
\ \\
One of the most prominent examples for a nonlinearity is the \textit{rectified linear unit} ReLU:
\begin{align}
\text{ReLU}(x):=\max(x,0)\label{ReLU}
\end{align}
This function has empirically been proven to be the most efficient choice in many cases, as its computation is faster than most other non-linear functions and it is less prone to plateaus in the training process of the network.\\
Note that ReLU is applied \textit{element-wise}, meaning that if a multi-channel feature map $f:\Z^2\to\R^K$ is given, ReLU is evaluated at each output channel $f_k(x),\,k=1,..,K$, individually. There are other forms of \textit{fiber-wise} nonlinearities, which will be touched on in section \ref{sectionsteercnns}.\\
\ \\

\noindent
\textbf{Pooling}\\
In CNNs, convolutional layers are often interspersed with so called \textit{pooling layers}, with one of the first examples being \cite{NIPS2012_c399862d}. In these layers, a pooling operation is performed on a given set of feature maps. The main goal of this operation is to cut out superfluous information, reducing data size by slightly reducing locational precision. This can be done because a CNN usually does not require the exact location of a feature to function properly. There are several different pooling operations, one of the most frequently used is \textit{max pooling}:
\begin{align}\label{gmaxpooling}
P:\Z^2\to \R;\hspace{1cm} Pf(x):=\underset{x'\in U(x)}{\max}f(x'),
\end{align} 
where $f$ is a feature map and $U(x)$ is some neighbourhood of $x$ in $\Z^2$, in this case usually being a small square with center $x$. The pooling operation thus works by finding the highest activation in said neighbourhood. Average pooling, which is another frequently used operation, would find the average of all activations in the pooling window.\\
To actually reduce the size of the feature map, pooling needs to be performed with a \textit{stride}, meaning that $P$ is not evaluated on every point in the base space, but only on points of certain distance to each other. For example, a stride of 2 would mean that in (\ref{gmaxpooling}), $P$ is only evaluated at points in $2\Z^2=\{2x\,:\,x\in \Z^2 \}$, while the neighbourhoods $U(x)$ remain subsets of $\Z^2$.

\subsection{Group Equivariant CNNs}\label{sectiongcnn}

\subsubsection{Motivation}

We can interpret convolutional networks from section \ref{cnnbasics} as already being G-CNNs with $G$ the group of translations, $\Z^2$. Our convolution operation $[f\star\psi](x)$ basically returns the activation of $\psi$, transformed under the action of the group element $x$, i.e. a translation. HG{For a general G-CNN, one} would like to find a way to replace the underlying group of the network with some other group $G$, containing more transformations than just translations, for example rotations and reflections, and still have equivariance of convolution with respect to every element of $G$.\\
\subsubsection{Group Convolution, Feature Maps and Equivariance}\label{sectiongconv}
\begin{figure}[h!]
	\centering
	\def\svgscale{0.7}
	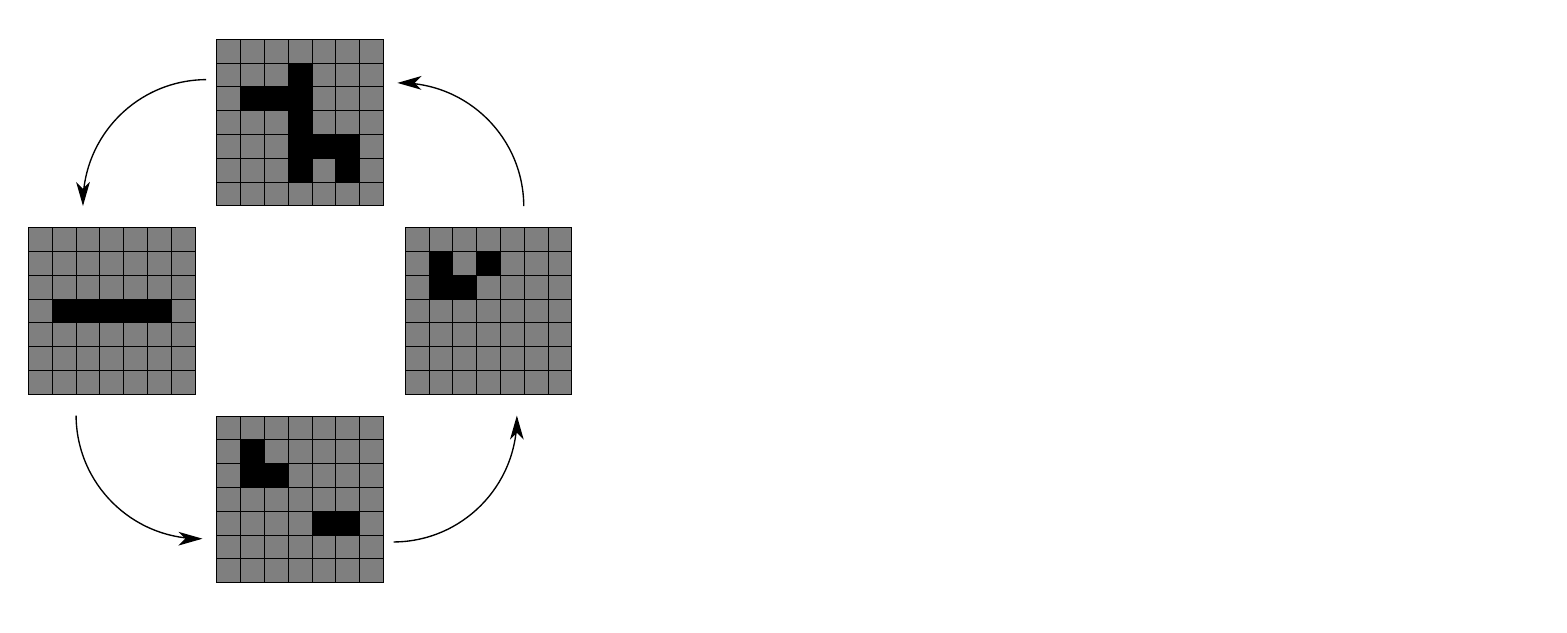
	\caption{A $p4$ feature map (left) and its rotation by $r$ (right).}
	\label{figure3}
\end{figure}

We first establish the action of the group $G$. 
The convolution operation from equation (\ref{convolution}) is modified in two steps. First, we consider the initial convolutional layer of the network, i.e. where the image $f:\Z^2\to \R$ is convolved with the first set of filters:
\begin{align}\label{gconv1}
f\star\psi :G\to \R; \hspace{1cm}[f\star\psi](g)=\sum_{y\in\Z^2}f(y)\psi(g^{-1}y)
\end{align}
Two things should be noted: First, $g\in G$ again transforms feature maps (or filters) via a group action on the domain of said maps:
\begin{align}\label{fmtransform}
[T_gf](x)=f(g^{-1}x); \;x\in\Z^2,\,g\in G.
\end{align}
\noindent
Specifically for $G=p4$ or $G=p4m$, $g\in G$ transforms $f(x)$ by moving its output from $x\in\Z^2$ to the pixel at $g^{-1}x$ under the actions defined in example \ref{explgroupaction}.\\
\ \\
\noindent
Secondly, it should be noted that the convolution operation $f\star \psi$ itself now yields a function on the group $G$ instead of $\Z^2$, thus requiring filters in the subsequent layer to also be functions on $G$, yielding yet another slightly different convolution operation:
\begin{align}\label{gconv2}
f\star\psi :G\to \R; \hspace{1cm}[f\star\psi](g)=\sum_{h\in G}f(h)\psi(g^{-1}h)
\end{align}
As one might notice, the transformation of feature maps and filters by $G$ is again slightly different:
\begin{align}\label{gfmtransform}
[T_gf](h)=f(g^{-1}h); \;h,g\in G.
\end{align}
Hence, the output of $f$ at a point $h\in G$ of the domain (which also is just $G$) is moved to the point $g^{-1}h$, which is obtained by the canonical action of $G$ on itself by its group operation.\\
\ \\
While it is easy to imagine feature maps $f:\Z^2\to\R$ simply as images sampled on a pixel grid, feature maps with base space $G$ might not be as intuitive. For $G=p4$, a feature map $f:G\to \R$ can be imagined as a graph with four patches, of which each corresponds to one of the four rotations $C_4=\{e,r,r^2,r^3\}$. Each pixel now has a rotational coordinate, corresponding to the patch in which it appears, and two translational coordinates specifying its position in the respective patch. These coordinates do also coincide with the semidirect product structure of $p4$, enabling us to write any element $g\in p4$ as a product $g=tr,\; t\in\Z^2,r\in C_4$.\\ 
\ \\
The rotation $r$ for instance now acts on this graph by moving each patch along the arrows indicated in figure \ref{figure3}, and by also rotating each of the patches themselves by 90 degrees, as is also shown in figure \ref{figure3}.

\noindent
The convolution operation of the first layer (\ref{gconv1}) and the convolution operations of the subsequent layers (\ref{gconv2}) are now not only equivariant to translations, but to all transformations from the group $G$, expanding translational equivariance by e.g. compositions of translations and rotations for $p4$ and by compositions of translations with rotations and reflections for $p4m$:
\begin{align}\label{gequivariance}
[[T_uf]\star\psi](g)=[T_u[f\star\psi]](g)\;\forall \,u,g\in G.
\end{align}
\noindent
For higher layer convolutions, this is derived in complete analogy to (\ref{transeq}), this time using the fact that $uG=G\;\forall u\in G$ and thus substituting $uh\in G$ for $h$, again keeping the overall sum the same. At the same point for first layer convolutions, we use an analogous argument $g\Z^2=\Z^2\;\forall g\in G$, as $G$ acts transitively on $\Z^2$, meaning that any point in the pixel grid can be reached from any other point by a transformation from $G$.

\subsubsection{Group Pooling and Equivariant Nonlinearities}\label{cosetpool}
\textbf{Equivariance to element-wise nonlinearities}\\
Element-wise nonlinearities $\nu:\R\to\R$ can be used in G-CNNs without restriction, as post-composing them with a feature map preserves the equivariance to pre-compositions with group transformations:\\
\ \\
Let $\nu:\R\to\R$ be an element-wise nonlinearity such as e.g. ReLU, and define the post-composition, i.e. the application of such a function to a feature map $f$: 
\begin{align*}
C_{\nu}(f(g))=[\nu\circ f](g)=\nu(f(g)).
\end{align*}
Let $T_g$ be the left transformation operator from (\ref{fmtransform}) or \eqref{gfmtransform}. As $T_g$ is realized by \textit{pre}-composition with $f$ and $C_{\nu}$ by \textit{post}-composition, these actions commute and we get
\begin{align}\label{nonlincommute}
C_{\nu}T_gf=\nu\circ[f\circ g^{-1}]=[\nu\circ f]\circ g^{-1}=T_gC_{\nu}f.
\end{align}

\noindent
\textbf{Group Pooling}\\
As in Regular CNNs, convolutional layers of a G-CNN are often followed by a pooling layer. As we no longer need to have the simple pixel grid $\Z^2$ as base space, we need to define the notions of pooling and stride for the more general case of the base space being a group $G$. To simplify the process, we split it into two steps, namely the pooling step, which is performed without stride, and the subsampling step which can then realize any notion of stride, if desired. In the first step, the max pooling operation for instance becomes
\begin{align}\label{gpooling}
P:G\to \R; \hspace{1cm} Pf(g):=\underset{x\in gU}{\max}\,f(x).
\end{align}

\begin{figure}[h!]
	\centering
	\footnotesize
	\def\svgscale{0.75}
	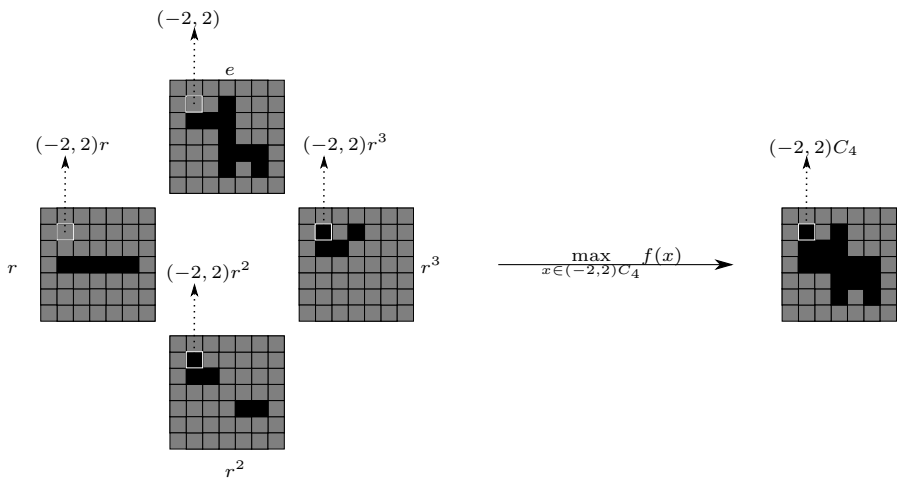
	\caption{Coset pooling by $H=C_4$ of a $p4$ feature map. A single coset ($(-2,2)C_4$) is highlighted.}
	\label{cosetpoolfigure}
\end{figure}

\noindent
Here, $gU:=\{gu\mid u\in U\}$ is some $g$-transformed neighbourhood $U$ of the identity element in G. In $\Z^2$, this would correspond to a square around the origin that is moved across the images by translations $t\in\Z^2$. This operation is now equivariant to $G$:
\begin{equation}\label{gpoolingeq}
\begin{split}
PT_hf(g)&=\underset{k\in gU}{\max}\,T_hf(k)\\
&=\underset{k\in gU}{\max}\,f(h^{-1}k)\\
&=\underset{hk\in gU}{\max}\,f(k)\\
&=\underset{k\in h^{-1}gU}{\max}\,f(k)\\
&=Pf(h^{-1}g)\\
&=T_hPf(g)
\end{split}
\end{equation}
\noindent
The arguments behind these equations are as follows: The first two equations are just the definitions of group pooling (\ref{gpooling}) and the left transformation of feature maps (\ref{gfmtransform}), respectively. In the third line, we substitute $k$ for $hk$ using the fact that taking the maximum of $f(h^{-1}k)$ over $k\in gU$ is the same as taking the maximum of $f(k)$ with $k$ such that $hk$ is in $gU$. In the fourth line we use that this is again the same as taking $k\in h^{-1}gU$, which can be seen by just multiplying both sides with $h^{-1}$. The last two lines are then obtained by just resubstitutuing the definitions of group pooling and $T_h$.\\
\ \\
\noindent
Any stride could now be realized in the second step by subsampling the pooled feature map over a subgroup $H\subseteq G$, i.e. evaluating just on points $h\in H$ instead of all $g\in G$. However, a feature map subsampled in this way would not anymore be equivariant to all of $G$, but only to $H$. As we wish to maintain equivariance to the whole group $G$, this form of stride is usually not used in practical $G$-CNN applications.\\
\ \\ 
Instead, to preserve $G$-equivariance throughout the network, one can use \textit{coset pooling} by choosing the pooling neighbourhood $U$ in the first step to be itself a subgroup $H$ of $G$. The resulting transformed pooling regions then are the non-overlapping \textit{cosets} of $H$ in $G$ which are either disjoint or equal for any $gH$ and $g'H$. Because of this, any element of each of the distinct cosets can then be chosen as a representative to subsample on. The resulting pooled feature map can then be interpreted as a map on the quotient space $G/H$, which can then be acted upon by $G$ similar to (\ref{gfmtransform}) by utilizing the general action of $G$ on its quotient spaces from example \ref{explgroupaction},  preserving equivariance as shown in (\ref{gpoolingeq}):
\begin{align}
T_{g'}f(gH)=f((g'^{-1}g)H),\; g'\in G, gH\in G/H
\end{align}
As an example of this, consider a $p4$-feature map that is pooled over the group of rotations, $C_4=\{e,r,r^2,r^3\}$. Visually (see figure \ref{cosetpoolfigure}), this equates to checking the four rotational outputs of each pixel coordinate $x\in\Z^2$ and choosing the one with the highest activation as representative for the coset $\{x,xr,xr^2,xr^3\}$. The resulting feature map is then of domain $p4/C_4=\Z^2$ and thus transforms in the same way as an input feature map in (\ref{fmtransform}).

\subsubsection{Implementation}
$G$-Convolution can be implemented rather easily, at least 
for so-called \textit{split} groups. By exploiting this property, we can just use a standard convolution routine with an expanded filter bank, which will be described shortly.\\
Recall that a group being \textit{split} means that any element $g\in G$ can be written as a product $g=ts$. For $p4$ and $p4m$, $t\in\Z^2$ would be a translation, and $s$ would be a transformation from the stabilizer group $H=C_4$ or $H=D_4$ that leaves the origin invariant, i.e. a rotation or roto-reflection around the origin. This, together with $T_tT_s=T_{ts}$ for the action of G allows us to rewrite the definition of $G$-convolution as follows:
\begin{align}\label{gconvsplit}
f\star\psi(g)=f\star\psi(ts)=\sum_{x\in X}f(x)T_t[T_s\psi(x)]
\end{align} 
with $X=\Z^2$ in layer one and $X=G$ in subsequent layers, thus allowing us to precompute the transformed filters $T_s\psi$ for all transformations of the stabilizer and then convolve them with the input using a fast planar convolution routine.\\
\ \\
The set of untransformed filters at some layer $l$ can be sorted in an array $F$ of shape $K^l\times K^{l-1}\times S^{l-1}\times n\times n$. Here, $K^{l-1}$ denotes the number of input channels, $K^l$ is the number of output channels i.e. the number of distinct filters, and $n\times n$ denotes the spatial extend of the filters. Furthermore, $S^{l-1}$ is the size of the stabilizer group, i.e. the number transformations of $G$ that fix the origin in the base space of the feature maps that $F$ is to be convolved with.\\
\ \\
Each transformation $T_s$ now ``acts'' on F by permuting the scalar entries of each of the $K^l\times K^{l-1}$ distinct ``filter blocks'' of shape $S^{l-1}\times n\times n$. If $S^l$ transformations are applied, this leads to an extended array of shape $K^l\times S^l\times K^{l-1}\times S^{l-1}\times n\times n$.\\
\ \\
The permutations themselves can be realized by implementing an invertible map $g$ which yields the group element corresponding to an index from the array of shape $S^{l-1}\times n\times n$, represented as matrices. For example, for $p4$ this map would be the following, as was described in example \ref{explgroupaction}, iv):
\begin{align}
g(s,u,v)=\begin{pmatrix}
\cos(\frac{s\pi}{2})&-\sin(\frac{s\pi}{2})&u\\
\sin(\frac{s\pi}{2})& \cos(\frac{s\pi}{2})&v\\
0&0&1
\end{pmatrix}
\end{align}
We then set 
\begin{align}
F^+[i,s',j,s,u,v]=F[i,j,\bar{s},\bar{u},\bar{v}]
\end{align}
with
\begin{align}
(\bar{s},\bar{u},\bar{v})=g^{-1}(g(s',0,0)^{-1}g(s,u,v)).
\end{align}
To use $F^+$ in a planar convolution routine, we exploit the fact that $X$ in (\ref{gconvsplit}) involves a sum over the stabilizer, again allowing us to to rewrite the equation as
\begin{align}
\sum_{x\in X}f(x)T_t[T_s\psi(x)]=\sum_{x\in \Z^2}\sum_{k=1}^{S^{l-1}K^{l-1}}f_k(x)T_t[T_s\psi_k(x)].
\end{align}
We can now reshape $F^+$ into an array of shape $S^lK^l\times S^{l-1}K^{l-1}\times n\times n$ which can then be applied to similarly reshaped feature maps in a planar convolution routine.

\subsection{Steerable CNNs}\label{sectionsteercnns}
\subsubsection{Motivation}
It was shown how G-CNNs achieve group equivariance by expanding the domain of the feature maps and filters. The magnitude of the expansion depends on the size of the stabilizer group of the origin. Thus, larger groups lead to larger expansions, which in turn lead to a proportionally increasing computing cost.\\
Steerable CNNs are a generalization of G-CNNs which achieve equivariance by defining filter banks as \textit{intertwiners}, which are morphisms between \textit{group representations}, through which feature spaces become \textit{G-steerable}. For a  classical, unrestricted filter bank to be an intertwiner, it needs to satisfy an \textit{equivariance constraint}, which depends on the representations that it is meant to intertwine. Thus, while G-CNNs achieve equivariance by expanding arbitrary filter banks, Steerable CNNs achieve it by restricting the space of available filters, thus decoupling the required computational power from the size of the group.
\subsubsection{Feature Spaces, Fibers and Steerability}\label{sectionsteerability}
\begin{figure}[h!]
	\centering
	
	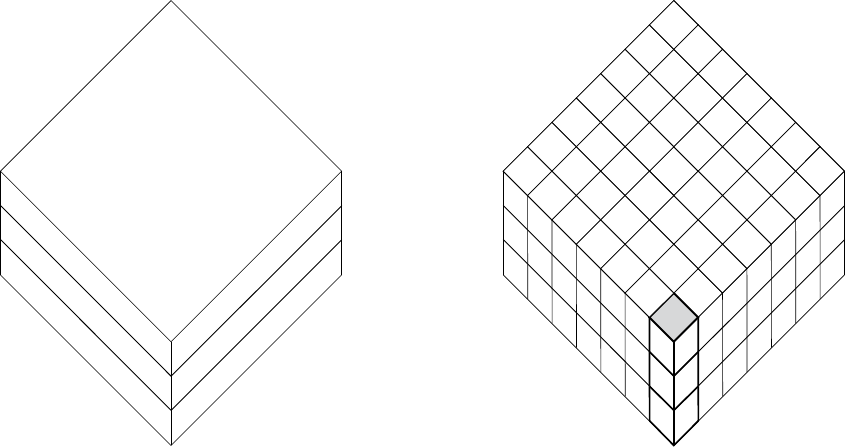
	\caption{The decomposition of $\F$ into stacks of feature maps (left), and into fibers (right). A single fiber is highlighted.}
	\label{figure5}
\end{figure}

We once again consider 2D signals $f:\Z^2\to\R^K$ with $K$ channels. These can be added, as well as multiplied by scalars and therefore form a vector space, often also called \textit{feature space}, which we will denote by $\mathcal{F}_l$. The layer index $l$ will often be omitted for simplicity.\\
\ \\ 
Given a group $G$ acting on $\Z^2$, we are able to transform signals $f\in \mathcal{F}_0$:
\begin{align}\label{pi0}
[\pi_0(g)f](x)=f(g^{-1}x).
\end{align}
\noindent
$\pi_0(g):\mathcal{F}_0\to \mathcal{F}_0$ is then a linear map $\forall g\in G$  that furthermore satisfies
\begin{align}\label{pi02}
\pi_0(gh) = \pi_0(g)\pi_0(h)\,\forall g,h \in G,
\end{align}
yielding a group homomorphism $\pi_0:G\to GL(\F_0)$.
A vector space such as $\mathcal{F}_0$, together with a map such as $\pi_0$ that satisfies (\ref{pi02}) fulfills the conditions of a group representation defined in definition \ref{defrepresentation}. It shall be denoted by $(\mathcal{F}_0,\pi_0)$. Oftentimes we will just call $\pi_0$ (or $\pi_l$) a representation, if the nature of $\mathcal{F}_0$ ($\F_l$) is clear. As will be explained later, the representations $\pi_l$ transforming higher layer feature spaces $\mathcal{F}_l$ might look slightly different than in (\ref{pi0}). Nevertheless, they will always satisfy (\ref{pi02}).\\ 
\ \\
While in most other deep learning publications some $f\in \mathcal{F}$ is usually considered as a stack of $K$ feature maps $f_k:\Z^2\to \R,\,k=1,..,K$, it is useful for the theory of steerable CNNs to consider another decomposition: Instead of splitting the feature space ``horizontally'' into one-dimensional planar feature maps, it can be decomposed into \textit{fibers} $\F_x$, one of which being located at each ``base point'' $x\in\Z^2$. Each fiber consists of the $K$-dimensional vector space representing all channels at the given position. $f\in \mathcal{F}$ is therefore comprised of feature vectors $f(x)\in \F_x$. See figure \ref{figure5} for an illustration.\\

\noindent
Let $\mathcal{F},\mathcal{F}'$ be two feature spaces, and let $\pi:G\to GL(\F)$ be a group homomorphism, such that $(\mathcal{F},\pi)$ is a group representation.
Let furthermore $\Phi$ describe a convolutional layer of a network, i.e.
\begin{align}
\Phi :\F\to\F';\hspace{1cm}\Phi(f)=f\star\Psi\;\forall f\in\F
\end{align}
for some filter bank $\Psi:\Z^2\to\R^K$.\\
\ \\
We now say $\F'$ is \textit{steerable} w.r.t. $G$, if there exists another function $\pi':G\to GL(\F')$ transforming $\mathcal{F}'$ such that
\begin{align}
\Phi\pi(g)f=\pi'(g)\Phi f\;\forall g\in G,
\end{align}
meaning that transforming the input by $\pi(g)$ yields the same result under $\Phi$ as transforming the output of $\Phi$ by $\pi'(g)$.
The following equation shows that this condition implies that $\pi':G\to GL(\F')$ is also a group homomorphism, and thus  $(\mathcal{F}',\pi')$ is also a group representation:
\begin{align}\label{pighom}
\pi'(gh)\Phi f= \Phi\pi(gh)f=\Phi\pi(g)\pi(h)f=\pi'(g)\Phi\pi(h)f=\pi'(g)\pi'(h)\Phi f.
\end{align}
Note that \eqref{pighom} only implies the desired property of $\pi'$ for the span of the image of $\Phi$. However, this is enough for our purposes, as any features in $\F'$ that are transformed by $\pi'$ result from applying a convolutional layer, i.e. lie in said image.

\subsubsection{The Equivariance Constraint on Filter Banks}
\begin{figure}[h!]
	\centering
	\def\svgscale{0.7}
	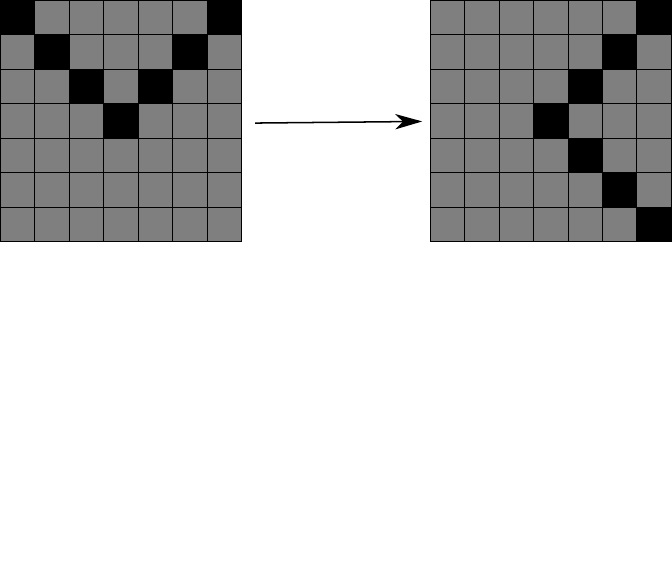
	\caption{A $H$-equivariant filter bank $\Psi$. Here, $\rho$ represents the rotation $r$ by cyclicly permuting the channels in each fiber. For $\Psi$ to be equivariant, this diagram needs to commute for any element of the considered group. Note that in this figure, we have $K=1$ and $K'=4$.}
	\label{figure6}
\end{figure}

Filterbanks are arrays of shape $K'\times K\times s\times s$, where $K'$ denotes the number of output channels, i.e. the number of distinct filters that are to be convolved with the input. Each of those $K'$ filters then has the shape $K\times s\times s$, where $K$ is the number of input channels, and $s$ denotes the spatial extent, i.e. the support/non-zero domain of the filter, usually being $3\times 3$ or $5\times 5$, though other sizes are possible.\\
\ \\
Assuming inductively that such a filter bank $\Psi$ is applied to a steerable feature space $(\mathcal{F},\pi)$, we need it to be an $H$\textit{-Intertwiner} between $\pi$ \textit{and} $\rho$, i.e. to satisfy the \textit{equivariance constraint} with respect to $H$, in order for the output of the convolution to be steerable:
\begin{align}\label{equivarianceconstraint}
\rho(h)\Psi=\Psi\pi(h)\,\forall h\in H.
\end{align}
This means that $\Psi$ applied to a feature map that was transformed by $\pi(h)$ has to yield the same result as applying a \textit{fiber representation} $\rho$ to the fiber that results from applying $\Psi$ to the untransformed image, as illustrated in figure \ref{figure6}.\\

\begin{figure}[h!]
	\centering
	\def\svgscale{0.7}
	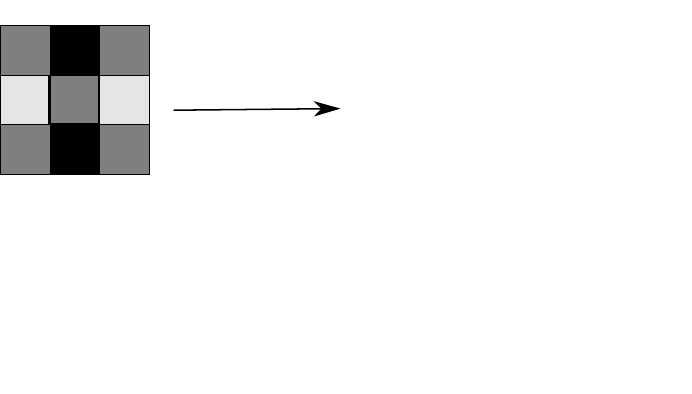
	\caption{Elements of the stabilizer group $H$ (top) do not shift patterns out of the receptive fields of fibers, while translations (bottom) do, thus making full $G$-equivariance of filter banks impossible.}
	\label{figure7}
\end{figure}

\noindent Two things should be noted at this point:\\
\ \\
Firstly, $\rho$ does not act on a feature space (or on a filter, equivalently) as $\pi_0$ does in \ref{pi0}, but on individual fibers $\mathcal{F}'_x\in\R^{K'}$. Formally, a fiber representation is thus just a $K'$-dimensional representation $(\R^{K'},\rho)$ of the stabilizer group $H$. In section \ref{sectioninduced}, it will be explained how a $G$-representation acting on a full feature space can be \textit{induced} from the $H$-representation $\rho$.\\ 
\ \\
Secondly, the equivariance constraint needs only to be fulfilled for all $h\in H$, thus excluding translations. This is because translations would be able to move patterns out of the receptive field of single fibers (see figure \ref{figure7}). This is relevant, as for the equivariance calculations we interpret the action of $\pi(h)$ on $\mathcal{F}$ for $h\in H$ as the equivalent action of $\pi(h^{-1})$ on the filters. Full $G$-equivariance will then be achieved by the induced representation. \\

\noindent
The equivariance constraint is linear in $\Psi$, meaning that any linear combination $\sum\lambda_i\Psi_i$ of intertwiners again satisfies (\ref{equivarianceconstraint}) and thus is  an intertwiner itself. We thus obtain a vector space of intertwiners, denoted $\text{Hom}_H(\pi,\rho)$.

\subsubsection{The Space of Intertwiners}\label{sectionintertwiners}
\begin{figure}[h!]
	\centering
	\def\svgscale{0.6}
	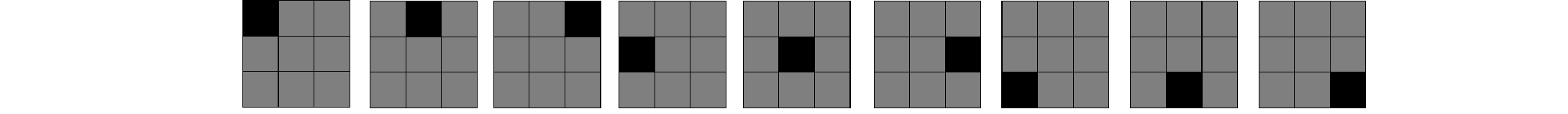
	\caption{The canonical Basis of $(\F_0,\pi_0)$.}
	\label{figure8}
\end{figure}

\begin{figure}[b!]
	\centering
	\def\svgscale{0.55}
	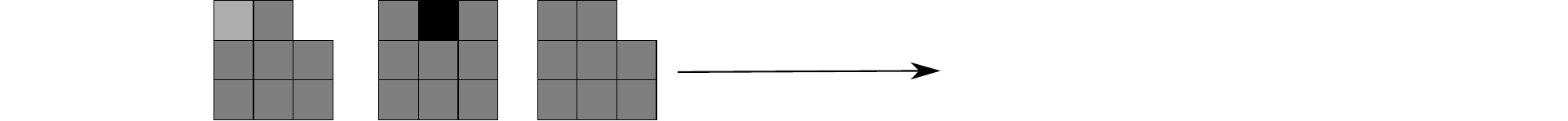
	\caption{$\pi_0(r)$ acting on a $\mathcal{C}$-linear combination.}
	\label{figure9}
\end{figure}

To better understand the construction of $\text{Hom}_H(\pi,\rho)$, we consider an example:
Let $\F_0$ be the space of $3\times 3$ filters with one channel, i.e. functions $\psi:\Z^2\to\R$ which return zero outside of the $3\times 3$ square around the origin. As these functions, just like feature maps, can be added and multiplied by scalars, $\F_0$ is a vector space of dimension 9 (or $Ks^2$ for arbitrary spatial extends and numbers of channels). $H$ can act on this space via $\pi_0$ from (\ref{pi0}) in the same way as it acts on feature maps:
\begin{align}
\pi_0(h)\psi(x)=\psi(h^{-1}x)
\end{align}
and, as mentioned in the previous section, transforming a patch from a feature map that $\psi$ is evaluated on by $h\in H$ is the same as transforming $\psi$ with $h^{-1}$. Staying with the example, ($\F_0,\pi_0$) is a 9-dimensional group representation of $H$.
The canonical Basis for this space is shown in figure \ref{figure8}.\\ Furthermore, an example for the transformation of a $\mathcal{C}$-linear combination of basis filters is depicted in figure \ref{figure9}.
\\
\noindent

\noindent
However, as explained in section \ref{sectionreptheory}, $\F_0$ can be decomposed into irreducible representations, i.e. subspaces $V\subseteq\F_0$, that are uniquely determined up to isomorphism (i.e. change of basis), and are $H$-invariant, meaning 
\begin{align}
\pi_0(h)V\subseteq V\;\forall h\in H.
\end{align}
For $G=p4m$ the irreducible decomposition of $(\F_0,\pi_0)$ for the stabilizer group $H=D_4$ is shown in table \ref{table4}.\\

\begin{table}
	\footnotesize
	\begin{center}
		
		\begin{tabular}{>{\centering}m{0.8cm}>{\centering}m{2.2cm}>{\centering}m{0.7cm}>{\centering}m{0.7cm}>{\centering}m{0.7cm}>{\centering}m{0.7cm}>{\centering}m{0.7cm}>{\centering}m{0.7cm}>{\centering}m{0.7cm}m{0.7cm}}
			Irrep&Basis Filters&$e$&$r$&$r^2$&$r^3$&$m$&$mr$&$mr^2$&$mr^3$\\
			\hline
			
			A1&\includegraphics[width=2.5cm,valign=c]{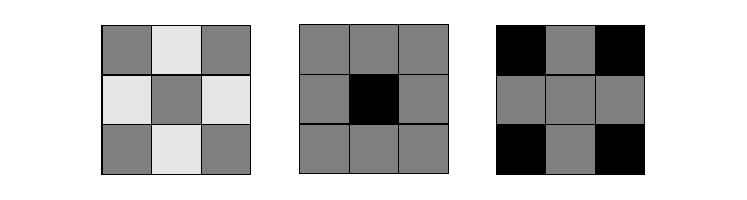}&$[1]$&$[1]$&$[1]$&$[1]$&$[1]$&$[1]$&$[1]$&$[1]$\\
			
			A2&\includegraphics[width=2.5cm,valign=c]{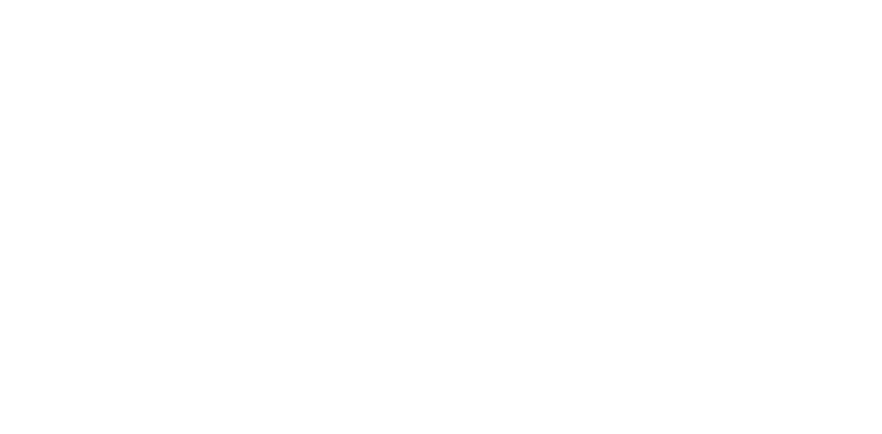}&$[1]$&$[1]$&$[1]$&$[1]$&$[-1]$&$[-1]$&$[-1]$&$[-1]$\\
			
			B1&\includegraphics[width=2.5cm,valign=c]{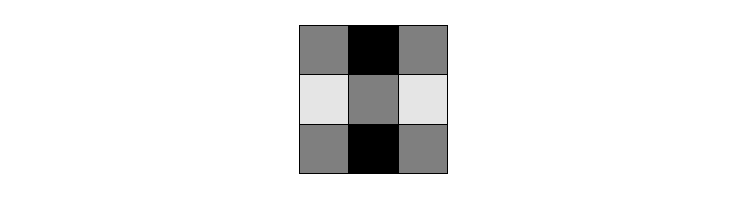}&$[1]$&$[-1]$&$[1]$&$[-1]$&$[1]$&$[-1]$&$[1]$&$[-1]$\\
			B2&\includegraphics[width=2.5cm,valign=c]{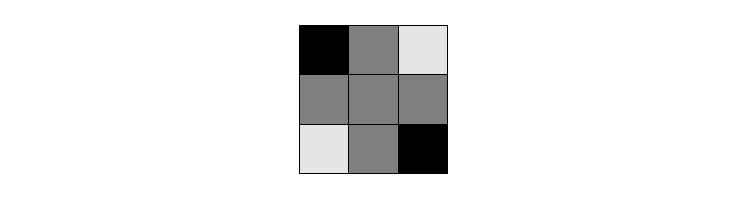}&$[1]$&$[-1]$&$[1]$&$[-1]$&$[-1]$&$[1]$&$[-1]$&$[1]$\\
			E&\includegraphics[width=2.5cm,valign=c]{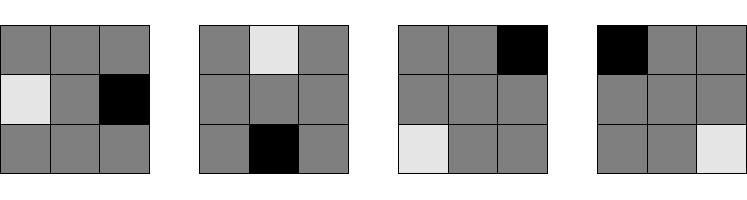}&$\begin{bmatrix}1&0\\0&1\end{bmatrix}$&$\begin{bmatrix}0&-1\\1&0\end{bmatrix}$&$\begin{bmatrix}-1&0\\0&-1\end{bmatrix}$&$\begin{bmatrix}0&1\\-1&0\end{bmatrix}$&$\begin{bmatrix}-1&0\\0&1\end{bmatrix}$&$\begin{bmatrix}0&1\\1&0\end{bmatrix}$&$\begin{bmatrix}1&0\\0&-1\end{bmatrix}$&$\begin{bmatrix}0&-1\\-1&0\end{bmatrix}$
		\end{tabular}\\
	\end{center}
	\caption{The irreducible decomposition of $(\F_0,\pi_0)$}
	\label{table4}
\end{table}

\begin{figure}[h!]
	\centering
	\footnotesize
	\def\svgscale{0.55}
	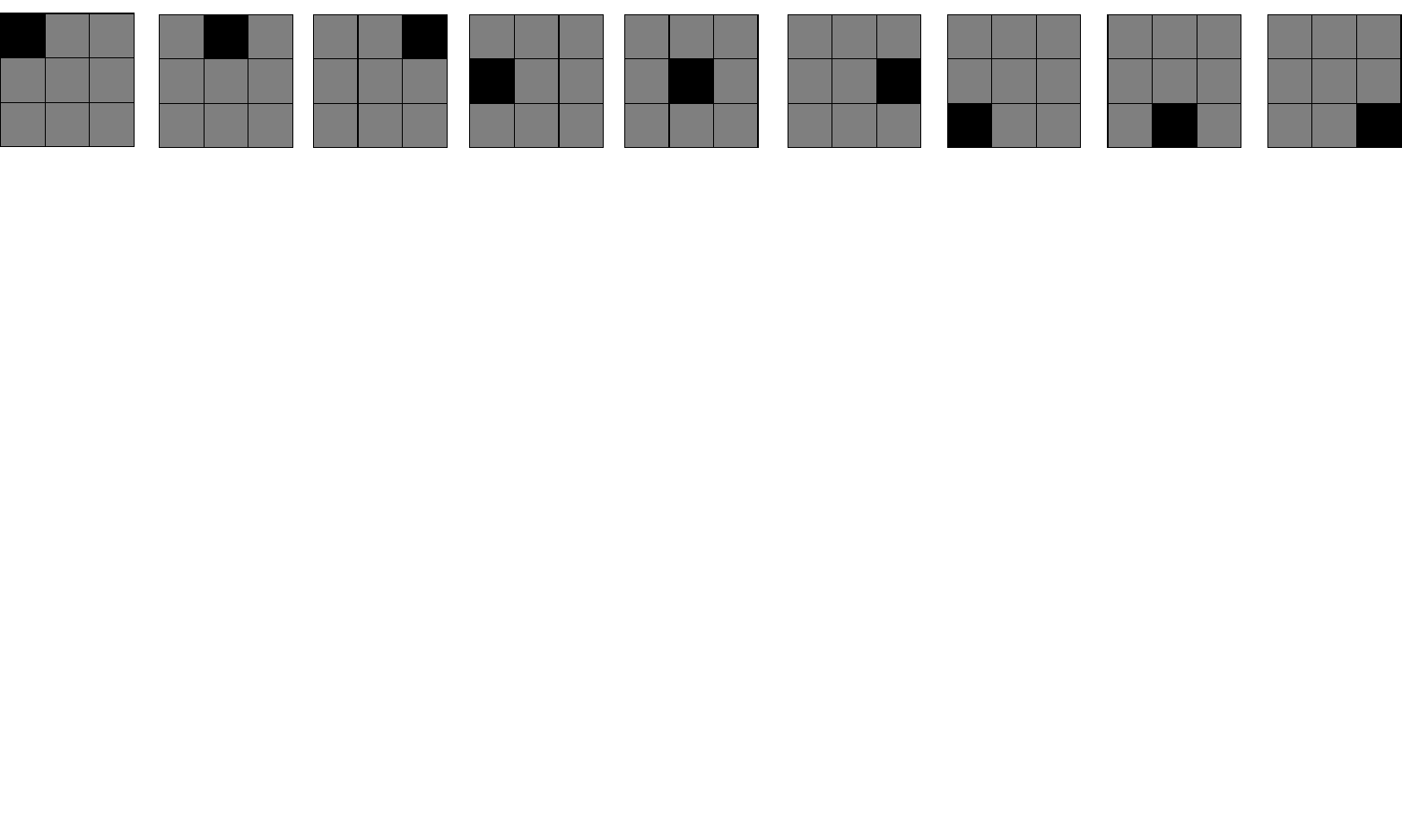
	\caption{The canonical basis $\mathcal{C}$ of $(\F_0,\pi_0)$ (top) is transformed by the reflection $\pi_0(m)$ (bottom). The invariant vectors are highlighted.}
	\label{figure10}
\end{figure}

\begin{figure}[b]
	\centering
	\def\svgscale{0.55}
	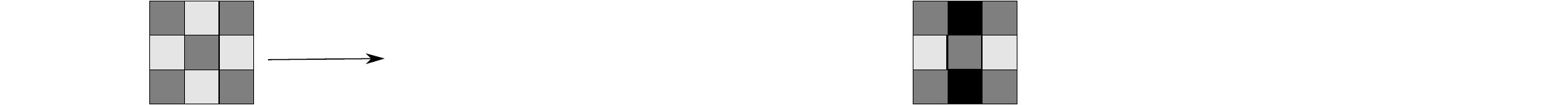
	\caption{A rotation acting trivially on an A1 basis element (left), and as [-1] on a B1 basis element (right).}
	\label{figure11}
\end{figure}

\noindent
As depicted, the group $H=D_4$ has five distinct irreps, which are characterized by the way in which $\pi_0(h)$ acts on them for different $h\in H$. For instance (see figure \ref{figure11}), $\pi_0(h)$ acts trivially on A1-filters, as rotating or reflecting has no effect on them, while $\pi_0(r)$ acts by multiplication by $[-1]$ on B1-filters.\\
\ \\
To obtain the irreducible decomposition of $\pi_0$, use a simplified  \textit{character formula}  (\cite{reeder2014notes},\cite{serre1977linear}):
\begin{align}
m_{\rho_i}(\pi_0)=\frac{1}{\lvert H\rvert}\sum_{h\in H}\chi_{\pi_0}(h)\chi_{\rho_i}(h).
\end{align}
The characters, which are just the traces of the representing matrices evaluated at each $h\in H$, can be obtained from table \ref{table4}. For other groups, such character tables are also available in literature. To determine the traces of $\pi_0(h)$ for $h\in H$, one can look at the canonical basis of $\F_0$ (figure \ref{figure8}) and how the individual vectors of it transform under $\pi_0$. The trace $\chi_{\pi_0}(h)$ of any representation matrix corresponding to $\pi_0(h)$ is then just the number of basis vectors of $\mathcal{C}$ that are left unchanged by its action. Thus, for instance, we receive $\chi_{\pi_0}(e)=9$, as any representation evaluated at the neutral element always acts as the identity matrix, and $\chi_{\pi_0}(m)=3$. The latter is visualized in figure \ref{figure10}.\\

\noindent
$\pi_0$ turns out to have type $(3,0,1,1,2)$ (see definition \ref{irreptype}), meaning that there are 3 copies of A1, i.e. 3 $H$-invariant one-dimensional subspaces with exactly the transformation properties of this irrep. Furthermore, there is one copy of each of B1 and B2, and two copies of the two-dimensional irrep E.\\

\noindent
Decomposing ($\F_0,\pi_0$) also makes $\pi_0(h)$ block diagonal for any $h\in H$, after applying a change of basis matrix $A$ that is constructed from the basis filters shown column two of table \ref{table4}. The same matrix $A$ block diagonalizes $\pi_0(h)$ for all $h\in H$ simultaneously and the elements on the diagonal are the irrep matrices shown in the third to last columns of table \ref{table4}, with their respective multiplicities. For instance, for $h=r$, we have:

\begin{align}
A\pi_0(r)A^{-1}=\text{block\_diag}([1],[1],[1],[-1],[-1],[0,-1;1,0],[0,-1;1,0])
\end{align}

\begin{figure}[h!]
	\centering
	
	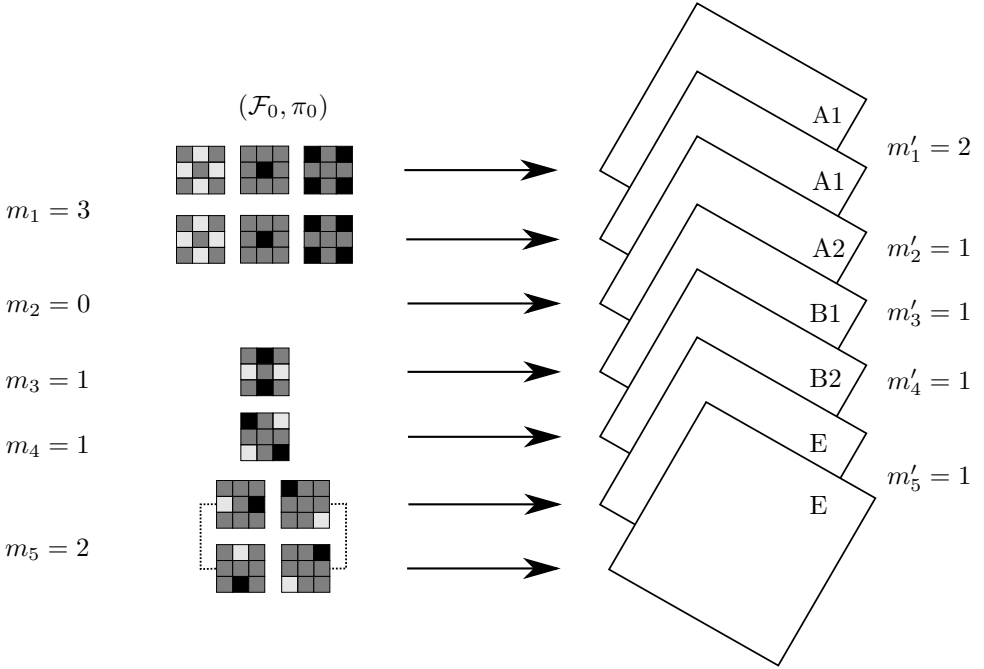
	\caption{The construction of an equivariant filter bank. The basis elements connected by the dotted lines must share a parameter to preserve equivariance. There are seven output channels.}
	\label{figure12}
\end{figure}

\noindent
To get an intuition for $\text{Hom}_H(\pi,\rho)$, we also need to look at the fiber representation $\rho$, which itself also has a decomposition into irreps and thus a type. In fact, $\rho$ would theoretically just be chosen by picking an arbitrary set of integers $(m_1,..,m_n)$ as multiplicities for the irreps ($n=5$ for $D_4$), as well as some basis $A\in\R^{K' \times K'}$. However, the choice of a basis is made obsolete later, when nonlinearities and \textit{capsules} are discussed. The number of output channels, i.e. the size of the fiber $\F_x$ that $\rho$ can act upon is also determined by the irreps:
\begin{align}
K'=\sum_{i=1}^{n}m_i\,\text{dim}\,\rho_i.
\end{align}
For the sake of this example, let $\rho$ be of type $(2,1,
1,1,1)$. Then, $K'=\sum_{i=1}^{5}m_i\,\text{dim}\,\rho_i=7$, thus $\rho$ acts on seven-dimensional fibers $\F_x\in\R^7$, for instance as
\begin{align}
\rho(r)=A^{-1}
\text{block\_diag}([1],[1],[1],[-1],[-1],[0,-1;1,0])
A,
\end{align}
i.e. a block diagonal matrix (after change of basis) with the corresponding matrices from column ``r'' of table \ref{table4} on the diagonal. The first channel now transforms by A1, the second by A2, and so on.\\
\ \\
As the number of output channels $K'$ equals the number of ``distinct filters'' that are convolved with the input, a filter bank $\Psi$ intertwining $\pi_0$ and $\rho$ must consist of seven filter units of shape $K\times s\times s$, i.e. $3\times 3$ in this case. Schur's Lemma (theorem \ref{schurslemma}) now determines which basis elements of $\F_0$ can be used to construct each of the individual filters.
According to the lemma, the intertwiner space of two irreps $\text{Hom}_H(\rho_i,\rho_j)$ is either zero, or one-dimensional iff $\rho_i$ and $\rho_j$ are isomorphic. It follows that each individual filter can only be constructed from the basis elements in $\F_0$ that transform under the same irrep as said filter's output channel.\\
\ \\
To finish the example, consider again the type of $\rho$. There are two A1-channels in the fiber, i.e. two distinct filters that need to be A1-equivariant. These filters can be constructed independently from one another, using the three basis elements in $\F_0$ that span the three A1-subspaces, yielding six parameters in total. There is one A2-channel, yet there are no basis elements in $\F_0$ that transform according to A2, thus this filter is simply zero, adding no additional dimensions. While the B1- and B2-filters are obtained in the same way as for A1, each yielding one independent parameter, there also is a copy of the two-dimensional irrep E in $\rho$. While this irrep therefore has two channels, these do not transform independently from one another, but are multiplied by two-dimensional matrices, which are given in the last row of table \ref{table4}. Thus, sets of two E2-basis filters in $\F_0$ (of which there also are two) have to share one parameter, yielding two parameters in total. An illustration of this construction is given by figure \ref{figure12}.\\

\noindent
The process demonstrated in this example can be generalized to intertwiner spaces $\text{Hom}_H(\pi,\rho)$ for arbitrary representations of respective types $(m_1,..,m_n)$ and $(m_1',..,m_n')$, where $\pi$ is the feature space representation that transforms feature maps and filters and $\rho$ is the fiber representation for the next layer. As mentioned before, $\pi$ might act on on the feature space (and thus on filters) in a slightly different way than $\pi_0$ (\ref{pi0}) used in this example. The reason for this will be explained in the next section. We receive the following formula for the dimension of of said intertwiner spaces, i.e. for the number of parameters required by any given layer:
\begin{align}
\text{ dim Hom}_H(\pi,\rho)=\sum m_i m_i'.
\end{align}

\subsubsection{Generating Steerable Feature Spaces}\label{sectioninduced}
What is left to be shown is how $H$-steerability of individual fibers $\F_x'$ leads to steerability of the whole feature space $\mathcal{F}'$ with respect to all of $G$. To derive this, we use the induced representation of $H$ on $G$ and we show that the transformation law that is imposed on an output space $\mathcal{F}'$. By convolving a transformed signal $\pi(g)f,\,f\in \mathcal{F}$ with an equivariant filter bank $\Psi\in \text{Hom}_H(\pi,\rho)$ we obtain  the formula for the induced representation. While the following paragraphs again give the computations the explicit groups $p4$ and $p4m$, the arguments used can be made in complete analogy for any other semidirect product $G=N\rtimes H$.\\
\ \\
Points $x\in\Z^2$ can be interpreted either as a point (e.g. when looked at as a pixel in the base space of feature maps), or as a translation. To make this interpretation explicit, we denote $x$ as $\bar{x}$ when we interpret it as a translation.\\
\ \\
\noindent
Translation equivariant convolution of $\Psi$ with $f$ can then be defined as
\begin{align}\label{convgeneral}
[ f\star\Psi](x)=\Psi[\pi(\bar{x})^{-1}f],
\end{align}
for translations $\bar{x}\in\Z^2$, with $\pi$ acting on $\F$ as described in (\ref{pi0}). \\
\ \\
We now utilize the semidirect product structure of $G$ (definition \ref{outersemidirectproduct}), which enables us to write any element $g\in G$ as a product $g=th$ where $t\in\Z^2$ is a translation and $h\in H$ is an element from the stabilizer group, i.e. some rotation from $C_4$ or rotation-flip from $D_4$ for $p4$ or $p4m$, respectively. It is useful to look at an explicit matrix representation of $G$. We can write any element of $G$ as
\begin{align}\label{matrixreps}
g=th=\begin{bmatrix}
I&T\\
0 &1
\end{bmatrix}
\begin{bmatrix}
R&0\\
0&1
\end{bmatrix}
=
\begin{bmatrix}
R&T\\0&1
\end{bmatrix}.
\end{align}
Here, $R$ is a matrix representation of $h\in H$, for instance a $2\times 2$ rotation or roto-reflection matrix for $p4$ or $p4m$, and T is the translation vector representing $t$. With this form of representation, a translation $\bar{x}\in\Z^2$ then amounts to\\
\begin{align}
\bar{x}=\begin{bmatrix}
I&x\\
0&1
\end{bmatrix}.
\end{align}
It is furthermore useful to make the difference between the action of $G$ on itself by group operation and the action of $G$ on $\Z^2$ visible, thus we will use $gh$ for $g,h\in G$ for the former and $g\cdot x$ for $g\in G, x\in\Z^2$ for the latter.\\
\ \\
\noindent
Applying $th$ to $f$ via $\pi$ before convolving with $\psi$ then yields:
\begin{equation}
\begin{split}
[\Psi\star [\pi(th)f]](x)&=\Psi\pi(\bar{x}^{-1})\pi(th)f\\
&=\Psi\pi(\bar{x}^{-1}th)f\\
&=\Psi\pi(hh^{-1}\bar{x}^{-1}th)f\\
&=\Psi\pi(h)\pi(h^{-1}\bar{x}^{-1}th)f\\
&=\rho(h)\Psi\pi(h^{-1}t^{-1}\bar{x}h)f\\
&=\rho(h)\Psi\pi(\overline{((th)^{-1}\cdot x)}^{-1})f\\
&=\rho(h)[\Psi\star f]((th)^{-1}\cdot x).
\end{split}
\end{equation}

\noindent
The justifications for these equations are as follows: In the first line, we use the definition of convolution (\ref{convgeneral}). In the second to fourth line, we use $hh^{-1}=e$, thus multiplying by one and the fact that $\pi$ is a group homomorphism, i.e. $\pi(gg')=\pi(g)\pi(g')\;\forall g, g'\in G$. The equivariance constraint (\ref{equivarianceconstraint}) is then used in the fifth line, and the argument of $\pi$ is inverted. In line six we use the fact that $h^{-1}t^{-1}\bar{x}h=\overline{((th)^{-1}\cdot x)}^{-1}$. This can be verified via the representation matrices presented in (\ref{matrixreps}). Finally, the definition of the convolution operation is applied backwards.\\
\ \\
\noindent
We can now define $\pi'$ by
\begin{align}\label{pi'}
[\pi'(th)f](x):=\rho(h)[f((th)^{-1}x)],
\end{align}
yielding $\Psi\star \pi(g)f=\pi'(g)\Psi\star f$, thus giving us the desired steerability of $\mathcal{F}'$ with respect to all of $G$.\\
\ \\
Comparing $\pi$ (\ref{pi0}) and $\pi'$ (\ref{pi'}), one notices that they only differ in the factor $\rho(r)$ which permutes the individual fibers after they have been moved from $x$ to $(th)^{-1}x$. This kind of action characterizes $\pi'$ as the aforementioned \textit{induced representation} of $\rho$ (of $H$) on $G$, often also denoted as $\text{Ind}_H^G(\rho)$. While there exists extensive knowledge about this concept, for instance in \cite{10.2307/2372309} and \cite{tomDieckrepheory}, for our purpose it suffices to know that induced representations transport fibers to new locations and then transform them via $\rho$, the representation induced from $\pi'$\\

\noindent
Furthermore, any representation of any subgroup of $G$ can induce a representation on $G$, allowing us to freely choose fiber representations when constructing filter banks. The representation $\pi_0$ from (\ref{pi0}) without any factor permuting the fibers can also be seen as being induced from the trivial representation, which acts as the identity matrix for all $h\in H$.\\
\ \\
It should also be noted that the content of section \ref{sectionintertwiners}, which is using  the decomposition of the trivially induced representation $\pi_0$, can also be applied to induced representations $\pi'=\text{Ind}_H^G(\rho)$. When investigating the action of such a representation $\pi'$ on basis filters to determine its trace, one  just has to include the factor of $\rho$, i.e. the fiber-permutation that is applied additionally.\\
\ \\
Further layers can now be added to the network by choosing a new representation $\rho'$ of $H$ to act on the fibers of the next output space and compute $\text{Hom}_H(\pi',\rho')$ for $\pi'$ restricted to $H$.\\

\subsubsection{Commutation with Nonlinearities via Capsules}\label{steercapsules}
We have now seen how the convolutional layers of steerable CNNs are built and how their equivariance to group transformations is achieved. In particular, it was shown that only the basis independent type of the input and output representations is relevant for the parameter count of a filterbank, i.e.
\begin{align*}
\text{dim }\text{Hom}_H(\pi,\rho)= \text{dim }\text{Hom}_H(\pi, A^{-1}\rho A)\;\forall A\in GL(\R^{\text{dim }\rho}).
\end{align*}
\noindent
However, when considering nonlinearities, the choice of the basis becomes relevant, which we shall demonstrate with a small example:\\
\ \\
Consider the classical ReLU function and let $g = (12)\in S_2$:
\[
\rho(g)=
\begin{pmatrix}
1&-1\\
0&1
\end{pmatrix}
\text{ and }
\rho'(g)=A^{-1}\rho(g) A=
\begin{pmatrix}
3&1\\
-4&-1
\end{pmatrix}\text{ for } A=\begin{pmatrix}
1&1\\2&1
\end{pmatrix}
\]
a change of basis.
$\rho(g)$, and thus $\rho'(g)$ correspond to a representation of $S_2$. 
But, e.g. for $x=(-2,5)^T$ and $x'=A^{-1}x=(7,-9)^T$ we have:
\begin{align*}
\text{ReLU}(\rho(g)(x))&=\text{ReLU}(\begin{pmatrix}
-7\\5
\end{pmatrix})=\begin{pmatrix}
0\\5
\end{pmatrix}\\
&\neq \text{ReLU}(\rho'(g)(x'))=\text{ReLU}(\begin{pmatrix}
12\\-19
\end{pmatrix})=\begin{pmatrix}
12\\0
\end{pmatrix}.
\end{align*}
Thus, different bases can lead to different results under non linear activation, even though the underlying type of the representation is the same. To efficiently tackle this problem, the concept of \textit{capsules} is introduced:\\
\ \\
A $\rho-$\textit{capsule} $(\R^K,\rho,\mathcal{B})$ is defined as a representation of the stabilizer group $(\R^K,\rho)$ with a fixed basis $\mathcal{B}$ of $\R^K$. By doing this, the representation matrices $\rho(h)$ also become fixed, which allows us us to \textit{realize} them as matrices, which will help in obtaining equivariance of nonlinearities, as will be explained in the following. In practice, capsules are typically held relatively low-dimensional, rarely exceeding the size of the stabilizer group, $\lvert H\rvert$. In the following a capsule will often just be denoted by $\rho$, omitting the specified Basis $\mathcal{B}$ and simply assuming that it is chosen in a way that yields a certain structure for the representation matrices.\\
\ \\
Just like a convolutional layer, a nonlinearity layer must be equivariant to the group $G$, which is realized by guaranteeing fiber-wise equivariance first and then ``inducing'' feature space equivariance afterwards.\\
\ \\
We define an element-wise nonlinearity $\nu:\R\to\R$, or a fiber-wise nonlinearity $\nu:\R^K\to\R^{K'}$ to be \textit{admissible} for an input representation $\rho$, iff there exists an output capsule $\rho$, such that
\begin{align}\label{eqconstraintnonlins}
\nu\rho(h)=\rho'(h)\nu\,\forall h\in H,
\end{align}
i.e. transforming input fibers by $\rho$ before applying $\nu$ is the same as transforming the nonlinearity's output by $\rho'$. We call $\rho'$ the \textit{post-activation capsule} corresponding to $\rho$ and denote it by $\text{Act}_{\nu}\rho$. \\
\ \\
\noindent
Given an admissible nonlinearity $\nu$ and corresponding pre- and post activation capsules $\rho$ and $\rho'=\text{Act}_{\nu}\rho$ such that (\ref{eqconstraintnonlins}) holds, feature space equivariance to the respective induced representation is derived as follows:
\begin{equation}
\begin{split}
\nu([\text{Ind}_H^G\rho](th)f)(x)&=\nu(\rho(r)f((th)^{-1}x))\\
&=\rho'(r)\nu(f((th)^{-1}x))\\
&=\rho'(r)\nu(f)((th)^{-1}x)\\
&=[\text{Ind}_H^G\rho'](th)\nu(f)(x)
\end{split}
\end{equation}

\noindent
Instead of building a layer by choosing multiplicities of irreps for a fiber representation, one now chooses multiplicities $m_i\geq 0, i=1,..,n$ corresponding to a set of predefined capsules $\rho_1,..,\rho_n$. The chosen capsules are then concatenated into a fiber of dimension $\sum m_i\,\text{dim}\,\rho_i$, which is then acted upon by $\rho=\bigoplus m_i\rho_i$, i.e. the block diagonal representation with $m_i$ copies of $\rho_i$ on the diagonal. Due to this block diagonal structure, capsules are \textit{disentangled}, i.e. channels of one capsule do not mix with channels belonging to other capsules during transformations.\\
\ \\
When it comes to finding explicit pairs of capsules and corresponding admissible nonlinearities, one needs not only to look at the irrep type of a capsule's underlying representation, but also at the individual form of the representation matrices $\rho(g)$ which, as stated, differs not only depending on the choice of irreps, but also on the choice of basis for the individual representation spaces. Below, we will name the most prominent choices of capsules. For a more extensive list, the reader is referred to \cite{weiler2019general}. \\
\ \\
First, any capsule $\rho$ which can be realized by permutation matrices will be compatible with any  element-wise nonlinearity. \textit{Element-wise} means that instead of acting on a whole feature vector $f(x)=(f_1(x),..,f_K(x))$, the nonlinearity transforms each element $f_i(x),\,i=1,..,K$, individually. The most prominent example for such a nonlinearity is the ReLU function (see \eqref{ReLU}). The most frequently used capsules that are realizable by permutation matrices are \textit{regular} capsules, i.e. feature vectors that transform according to the regular representation from example \ref{exquotrep}. While regular capsules have been shown to work very well in practice (\cite{cohen2016group},\cite{cohen2016steerable}), they have the drawback of leading to relatively high dimensional feature spaces, as each capsule has to span $\lvert H\rvert$ channels.\\
\ \\
A possible fix for the high dimensionality of regular capsules is given by \textit{quotient} capsules, in which the feature vectors transform according to the quotient representation (example \ref{exquotrep}) of $H$ by some (normal) subgroup $K$. Quotient capsules have the advantage of requiring $\frac{\lvert H\rvert}{\lvert K\rvert}$ instead of $\lvert H\rvert$ channels. If $K$ is normal, the features represented by such capsules become not just equivariant, but invariant to the symmetries of $K$:
\begin{align}
\rho_{\text{quot}}(k)e_{hK}=e_{khK}=e_{kKh}=e_{Kh}=e_{hK}.
\end{align}
Hence, performance can be improved by reducing parameter cost and feature size through the use of $H/K$ quotient capsules if the $K$-invariant patterns are prevalent in the input data. On the other hand, quotient capsules can also severely harm performance if the patterns are not found in the data or irrelevant to the learning task at hand. It should furthermore be noted, that quotient representations can also be obtained from non-normal subgroups. In that case, the features need not be necessarily invariant under $K$, but can perform differently. An example for this, as well as some visual intuition for quotient features in general, is given in appendix C of \cite{weiler2019general}.\\
\ \\
A second class of representation matrices are \textit{monomial} matrices, which have the same structure as permutation matrices, but also allow for entries of minus one instead of just one. Given a representation that is fully realized by monomial matrices, all \textit{concatenated} nonlinearities will be admissible. A concatenated nonlinearity is defined as evaluating an element-wise nonlinearity on an element $x$ and also on $-x$. For instance, the concatenated ReLU function is defined as
\begin{align}\label{CReLU}
\text{CReLU}(x):=(\text{ReLU}(x),\text{ReLU}(-x)).
\end{align}
Trivially, regular and quotient capsules will be compatible with concatenated nonlinearities. Also, many irreps can be realized (through a suitable basis) by monomial matrices, thus allowing for low dimensional irrep capsules. \\
\ \\
The third class of examples is given by \textit{orthogonal} matrices characterized by the fact that they preserve the norm (length) $\lvert x\rvert$ of vectors that they act upon. Capsules of this kind will be compatible with norm nonlinearities, which only act on the norm of any given feature vector (hence do not act element-wise), but not on its orientation. A norm nonlinearity can generally be written in the following form:
\begin{align}\label{normrelus}
\nu_{\text{norm}}:\R^K\to\R^K,\hspace{1cm}f(x)\mapsto\eta(\lvert f(x)\rvert)\frac{f(x)}{\lvert f(x)\rvert}
\end{align}
Here, $\eta:\R_{\geq 0}\to \R_{\geq 0}$ describes a non-linear function that is to act on the feature vectors norms. A prominent example of this would be \textit{Norm-ReLUs}, with
\begin{align}
\eta(\lvert ßf(x)\rvert)=\text{ReLU}(\lvert f(x)\rvert -b),
\end{align}
with $b$ being a learned bias. Amongst others, Norm-ReLUs were used in \cite{worrall2017harmonic}. The advantage of these nonlinearities is their broad compatibility, as any representation of any finite group can be made orthogonal, by choosing an orthogonal change of basis.
Further subclasses of norm nonlinearities, such as \textit{gated} and \textit{squashed} nonlinearities are mentioned in \cite{sabour2017dynamic} and \cite{weiler20183d}, respectively.
\subsubsection{A Remark on Pooling and Group Pooling in Steerable CNNs}\label{steerpool}
As far as the literature on steerable CNNs (\cite{cohen2016steerable},\cite{weiler20183d},\cite{weiler2019general}) is concerned, pooling operations are, if at all, applied fiber- or capsule-wise. This means that, instead of conventional pooling with stride, i.e. subsampling the feature map on a subgroup of $\Z^2$, thereby reducing equivariance of the network to that subgroup, each copy of a capsule located at each point $x\in\Z^2$ is searched for the maximal activation (in the case of max pooling), which is then set as the output of the pooling operation. Thus, pooling operations of this nature in steerable CNNs can be interpreted as nonlinearities, which were discussed in the previous section, and hence also need to satisfy an equivaraince constraint that is analogue to \ref{eqconstraintnonlins}.\\
\ \\
Let $P:\R^K\to \R$ be a pooling operation such as max pooling or average pooling, performed on a $K$-dimensional capsule $\rho$. For $G$-equivariance to be maintained, we need the pooling operation to be $H$-equivariant, i.e.
\begin{align}
P\rho(h)=\rho'(h)P\,\forall h\in H.
\end{align}
The choices of compatible capsules $\rho,\rho'$ is quite limited here, as they must not alter the numeric value of the feature vectors on which the pooling operation is performed, or the result of the pooling operation. Therefore, any input capsule $\rho$ that is to commute with $P$ has to be realized by permutation matrices, as taking the maximum (or average) value of a vector with permuted entries still yields the same result. Furthermore, $\rho'$ needs to be the trivial representation, i.e. $\rho'(h)=[1]\,\forall h\in H$ as otherwise equivariance could again be broken, for instance when the result is multiplied by [-1], as e.g. by the irrep A2 in table \ref{table4}. The most common choices for $\rho$ therefore are regular capsules, quotient capsules, as well as irrep capsules that are realized by permutation matrices.\\
\ \\
As an exception to the choice of $\rho'$ which was, to our knowledge, not yet discussed in the literature, one can implement an equivalent of coset pooling from \ref{cosetpool}. This will especially make sense taking into account that regular steerable CNNs are equivalent to $G$-CNNs in section \ref{sectionequivgsteer}. Suppose a subgroup $K\subseteq H$ is given, yielding a quotient space $H/K$. Setting the input capsule as the regular representation of $H$ and the output capsule as the quotient representation of $H/K$, we can define quotient pooling as follows:
\begin{align}\label{steercosetpool}
P_{\text{quot}}:\R^{\lvert H \rvert}\to\R^{\lvert H/K \rvert}, \;P_{\text{quot}}(f(x))=\underset{hK\in H/K}{\bigoplus}\underset{h'\in hK }{\max}f(x_{h'}).
\end{align}
This means that for each of the outputs $\underset{h'\in hN }{\max}f(x_{h'})$, $P_{\text{quot}}$ only looks at the coordinates $f(x_h)$ of $f(x)$ that correspond to the the elements of the respective coset that is being pooled over. This process is repeated $\lvert H/K \rvert$ times, i.e. for each coset. The individual results are then added into a $\lvert H/K \rvert$-dimensional vector that then transforms according to the quotient representation of $\lvert H/K \rvert$.\\

\subsubsection{Reduction of Parameters and Implementation}

By the use of steerable filter banks, the parameters of steerable CNNs are utilized several times more efficiently than the parameters of regular convolutional networks. As was shown in section \ref{sectionintertwiners}, an equivariant filter bank $\Psi$ intertwining representations $\pi$ and $\rho$ has a parameter cost of dim $\text{Hom}_H(\pi,\rho)=\sum m_im_i'$, depending on the irrep multiplicities $m_i$ and $m_i'$ of $\pi$ and $\rho$, respectively. A non-equivariant filter bank of the same size, i.e. same filter width $s\times s$ and same number of input/output channels $K$/$K'$ (which is also determined by the dimensions of $\pi$ and $\rho$), would instead have a parameter cost of $\text{dim } \pi \cdot \text{dim } \rho= s^2\cdot K\cdot K'$. \\
\ \\
Thus, the parameter efficiency of a steerable filter bank in comparison to a conventional filter bank is expressed as follows:
\begin{align}
\mu=\frac{\text{dim }\pi\cdot\text{dim }\rho}{\text{ dim Hom}_H(\pi,\rho)}.
\end{align}
For effective network architectures, this value usually lies around $\lvert H \rvert$, i.e. the size of the stabilizer group. For instance, we have $\mu=8$ for $H=D_4$, meaning that a $p4m$-steerable CNN's convolutional layer uses its parameters eight times more efficiently than a layer of the same size in a conventional CNN.
Efficiency is further increased by the fact that the representations $\pi$ and $\rho$ in each convolutional layer have a block-diagonal structure, i.e. consisting of direct sums of distinct, disentangled  and lower-dimensional capsules. Because of this, an intertwiner $\Psi$ between said representations will also have a certain block structure:\\
\ \\
For this, let $\pi=\pi_1\oplus ...\oplus \pi_p$ and $\rho=\rho_1\oplus ...\oplus\rho_r$. An intertwiner then is a matrix $\Psi$ of shape $K'\times Ks^2$, with $K'=\sum_{i=}^{r}\text{dim }\rho_i$ and $Ks^2=\sum_{i=1}^{p}\text{dim }\pi_i$, and has the following block structure:
\begin{align}
\begin{bmatrix}
h_{11}\in \text{Hom}_H(\pi_1,\rho_1)&\cdots &h_{p1}\in \text{Hom}_H(\pi_p,\rho_1)\\
\vdots &\ddots&\vdots \\
h_{1r}\in \text{Hom}_H(\pi_1,\rho_r)&\cdots &h_{pr}\in \text{Hom}_H(\pi_p,\rho_r)\\
\end{bmatrix}
\end{align}    
Here, each subblock $h_{ij}\in \text{Hom}_H(\pi_i,\rho_j)$ is itself an intertwiner between the capsules $\pi_i$ and $\rho_j$. In many implementations, the same capsule is used several, or even all the times, allowing to compute many or all of the $h_{ij}$ using the same intertwiner basis, thus drastically reducing computational cost. Ordering the individual capsules such that equivalent capsules are adjacent to each other then leads to superblocks $H_{ij}$ of shape $m_i\,\text{dim}\,\pi_i\times m_j\,\text{dim}\,\rho_j$, with $m_i,n_j$ being the respective multiplicities of the two capsules of $H_{ij}$. The superblock itself is then filled with the subblocks $h_{ij}$ of shape $\text{dim }\pi_i\times\text{ dim }\rho_j$.\\
\ \\
In practice, when a list of capsules $\rho_i$ and corresponding lists of post-activation capsules $\text{Act}_{\nu}\rho_i$ (depending on the nonlinearity) are given, the induced representations $\pi_i=\text{Ind}_H^G\text{Act}_{\nu}\rho_i$, as well as the bases for the intertwiner spaces $\text{Hom}_H(\pi_i,\rho_j)$ for all pairs $i,j$ are computed offline. The bases are stored as matrices $\psi_{ij}$ of shape $\text{dim }\pi_i\cdot\text{dim }\rho_j\times\text{ dim }\text{Hom}_H(\pi_i,\rho_j)$. After that, a list of input multiplicities $m_i$ and output multiplicities $n_j$ provided by the user is put into a parameter matrix $\Phi_{i,j}$ of shape $\text{dim }\text{Hom}_H(\pi_i,\rho_j)\times m_in_j$ and the aforementioned superblocks $H_{ij}$ are obtained by matrix multiplication of $\psi_{ij}\Phi_{i,j}$. After all superblocks are obtained, $\Psi$ is reshaped to $K'\times K\times s\times s$ and can then be convolved with the input.

\subsubsection{On the Equivalency of G-CNNs and Regular Steerable CNNs}\label{sectionequivgsteer}
\begin{figure}[h!]
	\centering
	\def\svgscale{0.8}
	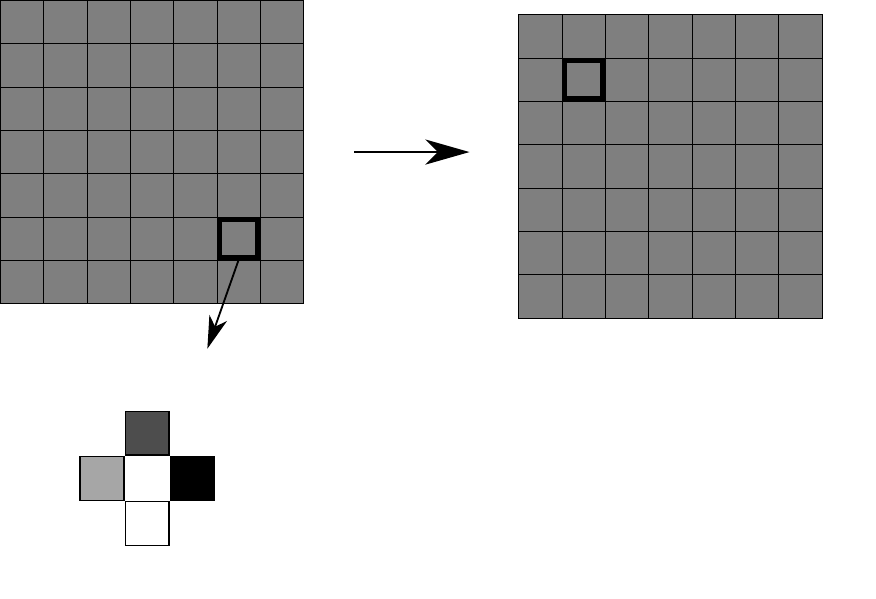
	\caption{The transformation of a single pixel $x\in\Z^2$ by the rotation $r\in C_4$ (top), and the permutation of that pixel's four rotational outputs (bottom). }
	\label{figure13}
\end{figure}

It was mentioned before that steerable CNNs are a direct generalizations of $G$-CNNs from section \ref{sectiongcnn}. To be precise, this means that $G$-CNNs are equivalent to steerable CNNs with regular capsules. To see this, first consider the feature maps of the respective architectures:
\begin{align}
f_G:G\to \R^{K_G},\hspace{1cm}f_{\text{steer}}:\Z^2\to\R^{K_s}.
\end{align}
In $G$-CNNs, all feature maps except the input image are functions on a group $G$, which in the cases that were considered here has $\Z^2$ as a normal subgroup. Equivariance is achieved by modifying the domain of the convolution operation, leading to $G$-convolution \eqref{gconv2} as described in section \ref{sectiongconv}. On the other hand, the feature maps of steerable CNNs keep the domain of $\Z^2$ in all layers and achieve equivariance by restricting the space of filter banks, as explained in detail in the sections \ref{sectionsteerability} to \ref{sectioninduced}.\\
\ \\
To understand the equivalence, we first look at how $f_G$ transforms under actions of $G$. For this, we again use the example of a $p4$-feature map (figure \ref{figure3}). A generalization to other groups is easily done.
The group $p4$ consists of unique tuples $x=(t,r)$ with $t\in\Z^2$ and $r\in C_4=\{e,r,r^2,r^3\}$. We can thus interpret a $p4$-feature map as a map on $\Z^2$, which returns four ``rotational coordinates'' at each point $t\in\Z^2$, corresponding to the elements $e,r,r^2,r^3$, as is depicted in figure \ref{figure13}, which gives a visualization of this slightly different, but equivalent interpretation of a $p4$-feature map. When being acted upon by another element $x'=(t',r')\in p4$, the $\Z^2$-pixel coordinate $t$ of $x$ and its rotational outputs are moved to $x'^{-1}t$. Simultaneously, the rotational coordinates are permuted by $r'$.\\  

\noindent
In steerable CNNs, all feature maps and filters are functions on $\Z^2$. The equivalent of the rotational outputs of $G$-CNNs is realized by the channels of the regular representation of $C_4$, which has a dimensionality of $\lvert C_4 \rvert=4$, thus consisting of one channel per element of $C_4$. As the transformation law of the regular representation on these four channels is defined by the group's action on itself by composition, they transform in exactly the same way as the rotational outputs of a $G$-CNN. The induced representation $\text{Ind}_{C_4}^{p4}\rho_{\text{reg}}$ of this regular representation lastly ensures that the respective pixel coordinates $t\in\Z^2$ transform in the same way in both cases, as can be checked by comparison of the equations \eqref{fmtransform} and \eqref{gfmtransform} with \eqref{pi0} and \eqref{pi'}. Hence, we receive the desired equivalence of $G$-CNNs and regular steerable CNNs. As stated, these arguments can be generalized to any semidirect product group $G'=N\rtimes H$ instead of $\Z^2\rtimes C_4$, as the regular representation behaves in the same way for any finite stabilizer group $H$. In any case, one channel of a $G$-feature map in a layer of a $G$-CNN equates to one regular capsule, and thus to $\lvert H \rvert$ channels in the equivalent regular steerable CNN.

\section{Steerable CNNs on $\Z^n\rtimes S_n$}\label{sectionsnapplications}
In this section, possible applications of the theory of equivariant networks to the symmetric group $S_n$ will be explored. As $G$-CNNs are equivalent to a special case of steerable CNNs, this chapter will stay in the more general language of the latter.\\
 \\
As in example \ref{explsn}, one can build semidirect product groups $\Z^n\rtimes S_n$, with the most practical use cases likely being $n=2$, for instance for 2D-images and $n=3$ for volumetric data, as for instance in \cite{weiler20183d}. One use case is given by interpretation of street scenes by one neural network based on the input of  two redundant camera sensors that are installed at nearby positions. The signal of these sensors can be considered to be equivalent, but not necessarily equal, as one camera sensor could be disturbed by dirt, but not the other. These networks will be very similar to the ones discussed before, with the difference that $S_n$ is used as the stabilizer group instead of $C_4$ or $D_4$.\\
  
\subsection{Steerable Feature Spaces and Filter Banks for $\Z^n\rtimes S_n$}
$\Z^n\rtimes S_n$ (see example \ref{explsn}) acts on functions on $\Z^n$ in a similar way to the product groups $p4$ and $p4m$ by compositions of translations and coordinate permutations. Hence, the theory of steerable CNNs from section \ref{sectionsteercnns} can be applied for $\Z^n\rtimes S_n$ in near complete analogy. In this section, we will nevertheless summarize the process of obtaining steerable feature spaces and filter banks. We also highlight the key differences, which lie in representation theory, such as in basis filters for intertwiner spaces, as these depend on the irreps of the underlying group. When discussing concrete examples, we will confine to $S_2$ and $S_3$, for visualization purposes. However, the general procedure of constructing steerable CNNs on $\Z^n\rtimes S_n$ for any $n\in\N$ becomes apparent. \\
\ \\
\textbf{Feature Spaces}\\
We again deal with feature spaces $\mathcal{F}_l$ of functions $f:\Z^n\to\R^{K_l}$ with $K_l$ output channels in each layer $l$, which transform by a representation $\pi_l$ analogously to \eqref{pi0} for $l=0$ and \eqref{pi'} in higher layers:
\begin{equation}\label{pi0sn}
[\pi_0(t,\sigma)f] (x)=f((t\sigma)^{-1}x)=f(\sigma^{-1}(x-t)),\,t\in\Z^n,\sigma\in S_n
\end{equation}
\begin{equation}\label{pilsn}
[\pi_{l+1}(t,\sigma)f](x)=\rho_{l}(\sigma)[f((t\sigma)^{-1}x)],\,t\in\Z^n,\sigma\in S_n.
\end{equation}  
In (\ref{pilsn}), $\rho_{l}$ is the fiber representation from the previous layer from which $\pi_{l+1}=\text{Ind}_{S_n}^{\Z^n\rtimes S_n}\rho_l$ is induced, and with respect to which filter banks $\Psi\in\text{Hom}_{S_n}(\pi_l,\rho_l)$ of layer $l$ must be equivariant:
\begin{align}
\rho_l(\sigma)\Psi=\Psi\pi_l(\sigma)\;\forall \sigma\in S_n.
\end{align}
\textbf{Equivariant Filter Banks}\\
The symmetric group of 2 elements has two distinct irreps: The trivial representation, id, and the sign representation sgn, both of them have dimension one. By restricting $\pi_0$ from (\ref{pi0sn}) to $S_2$ and letting it act on $3\times 3$ filters with one channel, we once again receive a nine-dimensional representation $(\F_0^{S_2},\pi_0)$. One now obtains the irrep decomposition of $\pi_0$ by applying the character formula for $\chi_{\text{id}}$ and $\chi_{\text{sgn}}$:
\begin{align}
m_{\pi_0}(\text{id})=\frac{1}{\lvert S_2\rvert}\sum_{\sigma\in S_2}\chi_{\pi_0}(\sigma)\chi_{\text{id}}(\sigma)=\frac{1}{2}(\chi_{\pi_0}(e)\chi_{\text{id}}(e)+\chi_{\pi_0}((12))\chi_{\text{id}}((12)))=6.
\end{align}
While $\chi_{\text{id}}(e)$ and $\chi_{\text{id}}((12))$ can be looked up from table \ref{table5}, it is useful for $\chi_{\pi_0}$ to consider the canonical basis vectors of  $(\F_0^{S_2},\pi_0)$ in figure \ref{figure8} and how $\pi_0(e)$ and $\pi_0((12))$ transform them. The representation matrix for the identity element is always the identity matrix for any representation, hence we receive $\chi_{\pi_0}(e)=9$. On the other hand, $\pi_0((12))$ swaps the coordinates of each pixel in each basis element, thus only leaving the third, fifth, and seventh filter invariant, leading to $\chi_{\pi_0}((12))=3$.\\
\ \\
Since there is only one other possible irrep, we immediately get $m_{\pi_0}(\text{sgn})=3$. With this, we have the full irreducible decomposition of $(\F_0^{S_2},\pi_0)$, which is depicted with an exemplary set of basis filters in table \ref{table5}.

\begin{table}
	\centering
	\begin{tabular}{c|c|c|c}
		Irrep&Basis Filters&$e$&$(12)$\\
		\hline
		&&&\\\
		id&\includegraphics[width=3cm,valign=c]{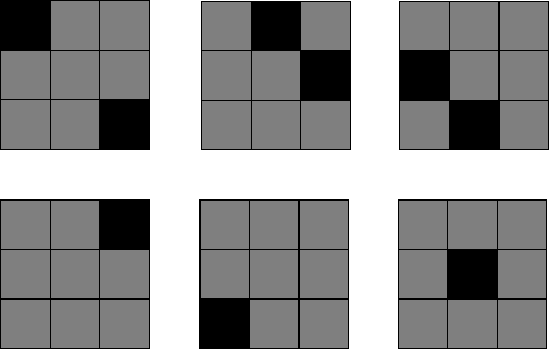}&$[1]$&$[1]$\\
		&&&\\
		$\text{sgn}$&\includegraphics[width=3cm,valign=c]{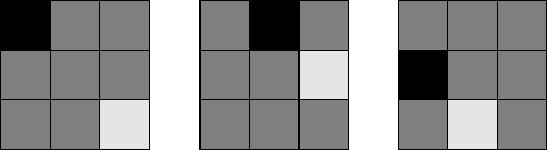}&$[1]$&$[-1]$\\
		
	\end{tabular}\\
	\caption{The irreducible decomposition of $(\F_0^{S_2},\pi_0)$ for $S_2$.}
	\label{table5}
\end{table}

\begin{figure}[h!]
	\centering
	\def\svgscale{0.75}
	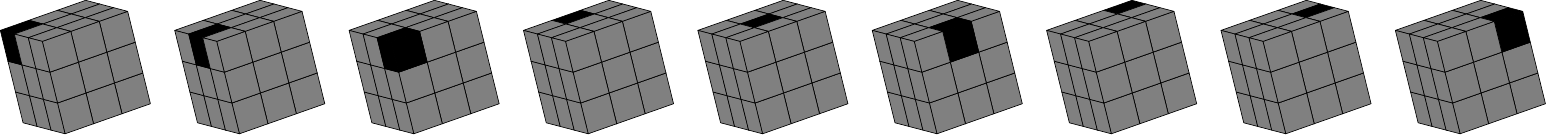
	\caption{Some canonical basis elements of $\mathcal{C}_{S_3}$ of $(\F_0^{S_3},\pi_0)$. As always, entries of one are black and entries of zero are grey.}
	\label{figure15}
\end{figure}

\noindent
For $S_3$ and $\Z^3$, we have one extra dimension for the base of feature spaces and filters, hence leading to an increase in dimensionality of the space of filters when looking at the $S_3$-equivalent of $(\F_0^{S_2},\pi_0)$. Instead of $3\times 3$ filters, we now have $3\times 3\times 3$ filters, leading to a representation $(\F_0^{S_3},\pi_0)$ of degree 27. The canonical basis for the space $\F_0^{S_3}$ is depicted in figure \ref{figure15}.\\
\ \\
We next determine the irreducible decomposition of $\pi_0$ in $\F_0^{S_3}$. While the characters of the three irreducible representations of $S_3$ are given in table \ref{table6}, it might look tedious at first to determine the character of $\pi_0$. However, this becomes quite easy if one again looks at which basis elements from $\mathcal{C}_{S_3}$ (figure \ref{figure15}) are left invariant by each group element of $S_3$. As always, $\pi_0(e)$ leaves every element invariant, thus yielding $\chi_{\pi_0}(e)=27$. For the other elements, it is helpful to identify each basis vector with the coordinates of its non-zero entry (i.e. the black cube in each filter), e.g. the first vector with $(-1,1,1)$,the second with $(0,1,1)$ and the fourth with $(-1,1,0)$. \\
\ \\
Any of the transpositions, i.e. 2-cycles $(12),(23),(13)\in S_3$ fix all vectors of which the two non-zero coordinates that are permuted are equal, e.g. $\{(x_1,x_1,x_2)\mid x_1,x_2\in \{-1,0,1\}\}$ for $(12)$. Thus, each transposition fixes 9 elements, so we have $\chi_{\pi_0}((12)=\chi_{\pi_0}((23)=\chi_{\pi_0}((13)=9$. The same argument can be used for the two 3-cycles. Each of them fixes the same 3 vectors of which all three non-zero coordinates are equal, hence we have $\chi_{\pi_0}((123))=\chi_{\pi_0}((132))=3$.

\begin{figure}[h!]
	\centering
	\def\svgscale{0.65}
	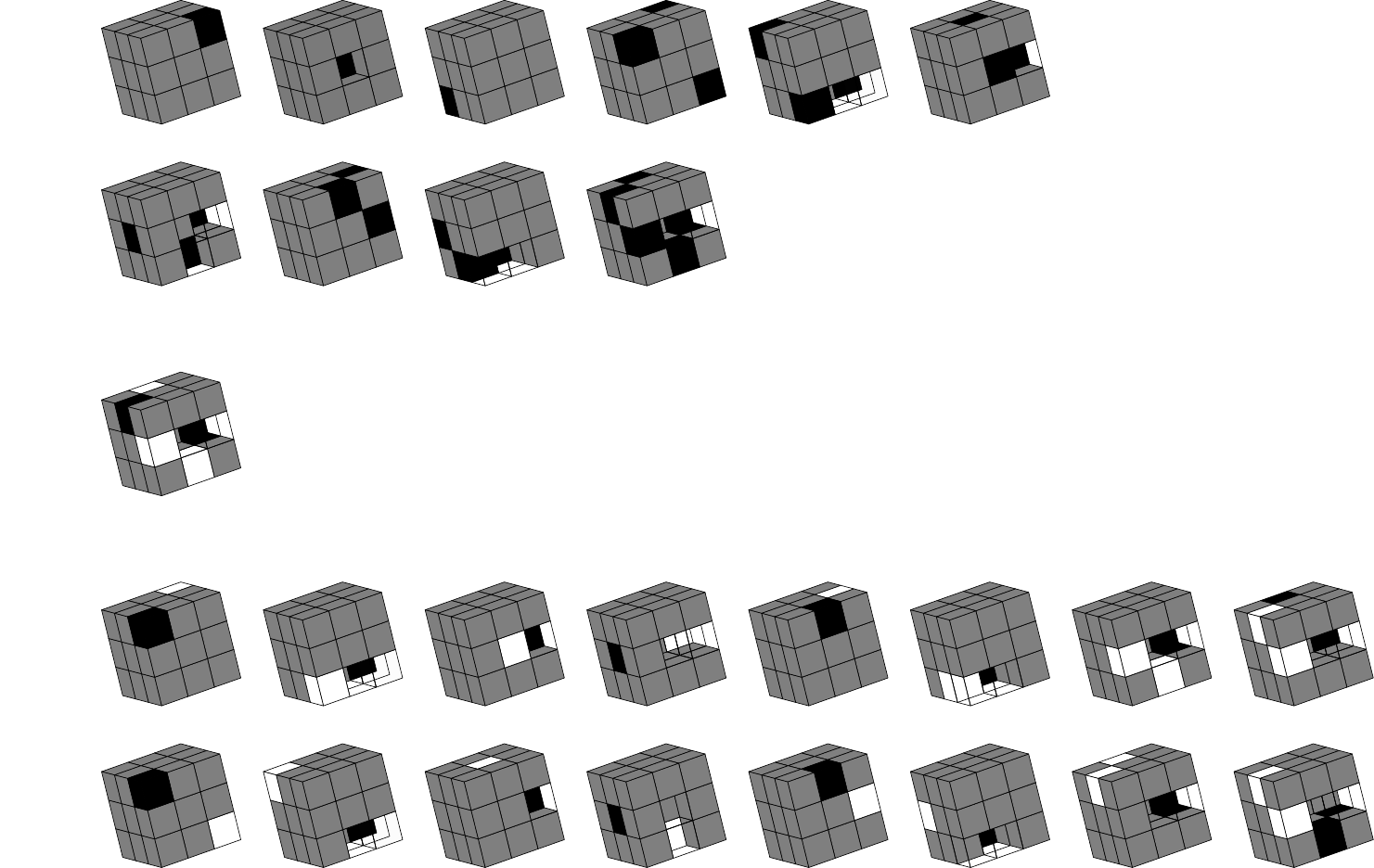
	\caption{The basis vectors for the irrep decomposition of $(\F_0^{S_3},\pi_0)$. The reader may verify that these vectors transform under $\pi_0$ by the respective matrices given in table \ref{table6}. Again, grey and transparent entries count as zeros, black entries as one and white entries as minus one.}
	\label{figure16}
\end{figure}

\ \\
\noindent
Plugging these results into the character formula now yields

\begin{equation}
\begin{gathered}
m_{\pi_0}(\text{id})=\frac{1}{\lvert S_3\rvert}\sum_{\sigma\in S_3}\chi_{\pi_0}(\sigma)\chi_{\text{id}}(\sigma)=\frac{1}{6}(27+3\cdot 9+2\cdot 3)=10,\\
m_{\pi_0}(\text{sgn})=\frac{1}{\lvert S_3\rvert}\sum_{\sigma\in S_3}\chi_{\pi_0}(\sigma)\chi_{\text{sgn}}(\sigma)=\frac{1}{6}(27+3\cdot (9\cdot (-1))+2\cdot 3)=1,\\
m_{\pi_0}(V_s)=\frac{1}{\lvert S_3\rvert}\sum_{\sigma\in S_3}\chi_{\pi_0}(\sigma)\chi_{V_s}(\sigma)=\frac{1}{6}(27\cdot 2 +3\cdot (9\cdot 0)+2\cdot (3\cdot (-1)))=8.\\
\end{gathered}
\end{equation}
Thus, the type (i.e. the multiplicities of irreps) of $\pi_0$ is $(10,1,8)$ for the trivial representation id, the sign representation sgn and the two-dimensional standard representation $V_s$, respectively. The basis of this irreducible decomposition is depicted in figure \ref{figure16}. As one can verify, these filters transform under $\pi_0(\sigma)$ for $\sigma\in S_3$ by multiplication with the representation matrices for the respective group element, which are given in table \ref{table6}.\\

\noindent
Equivariant filter banks and steerable feature spaces are now obtained in analogy to the sections \ref{sectionintertwiners} and \ref{sectioninduced}. Given some list of output multiplicities $(m_{\text{id}},m_{\text{sgn}},m_{V_s})$ (or just $(m_{\text{id}},m_{\text{sgn}})$ for $S_2$) of some fiber representation $\rho$, the filter bank $\Psi\in\text{Hom}_H(\pi_0,\rho)$ is constructed by linearly combining the irrep basis filters from figure \ref{figure16} for $S_3$ or from table \ref{table5} for $S_2$ in the same way as is depicted (for $p4m$) in figure \ref{figure12}, i.e. by only combining filters that correspond to the irrep that each of the respective output channels is transformed by.\\
\begin{table}
	\centering
	\renewcommand{\arraystretch}{1.2}
	\begin{tabular}{c|c|c|c|c|c|c}
		
		Irrep&$e$&$(12)$&$(13)$&$(23)$&$(123)$&$(132)$\\
		\hline
		id&$[1]$&$[1]$&$[1]$&$[1]$&$[1]$&$[1]$\\	
		$\text{sgn}$&$[1]$&$[-1]$&$[-1]$&$[-1]$&$[1]$&$[1]$\\
		$V_s$&$\begin{bmatrix}1&0\\0&1\end{bmatrix}$&$\begin{bmatrix}1&0\\-1&-1\end{bmatrix}$&$\begin{bmatrix}-1&1\\0&-1\end{bmatrix}$&$\begin{bmatrix}0&1\\1&0\end{bmatrix}$&$\begin{bmatrix}-1&1\\-1&0\end{bmatrix}$&$\begin{bmatrix}0&1\\-1&-1\end{bmatrix}$		
	\end{tabular}
	\caption{The representation matrices of $\pi_0$ for the basis filters of the irrep decomposition of $(\F_0^{S_3},\pi_0)$.}
	\label{table6}
\end{table}\\
\noindent
\subsection{Capsules, Nonlinearities and Pooling for $\Z^n\rtimes S^n$}
As was explained in the sections \ref{steercapsules} and \ref{steerpool}, certain requirements have to be fulfilled for a layer (realized by a concatenation of capsules as before) to commute with fiber-wise nonlinearities and pooling. A layer is thus not built by choosing multiplicities for the irreps directly, but by choosing copies of capsules to be concatenated into a fiber. We now list some relevant capsules for steerable CNNs with $H=S_2$ or $H=S_3$\\
\ \\
As for any finite group, we can look at regular capsules which transform under the regular representation $\rho_{\text{reg}}$. For $S_2$, this is the two-dimensional representation of type $(1,1)$ and for $S_3$, we have the four-dimensional representation of type $(1,1,2)$. As all regular capsules, these are realized by permutation matrices and thus commute with all nonlinearities discussed in \ref{steercapsules}, including ReLU. Furthermore, fiber-wise max-pooling can be applied. To obtain quotient capsules (which have the same compatibility properties as regular capsules), subgroups are needed. Of these there are none except the group itself and $\{e\}$ for $S_2$, which would lead to a trivial capsule and a regular capsule, respectively. For $S_3$, we have (besides the aforementioned ones) one normal subgroup, which is the \textit{alternating} group containing all even permutations, $A_3=\{e,(123),(132)\}$. With this, we have $S_3/A_3=S_2$, hence this quotient capsule would behave like a regular capsule of $S_2$. $S_3$ furthermore has $S_2$ as non-normal subgroup, allowing for $S_3/S_2$-quotient capsules. The two subgroups can also be used to implement coset pooling as described in \ref{steercosetpool}.

\section{Conclusion}
In this review article, we have seen several relevant approaches to achieve invariant representations in machine learning. After establishing the mathematical preliminaries from group theory and representation theory, we discussed translation equivariant convolutional networks. We then saw how group equivariant neural networks generalize CNNs and thus allow for representations that are invariant to groups of transformations that are more general than just translations. This concept was then generalized once more by introducing steerable CNNs, which by the use of group representations allow for different, more nuanced ways to express equivariance to a group.\\
We furthermore presented an application of the theory to the symmetric group resulting in a steerable CNN architecture for the symmetric group.

\vspace{2ex}

\noindent\textbf{Acknowledgement.} Helpful discussions with Matthias Rottmann are gratefully acknowledged.

\bibstyle{plain}
\bibliography{sn-bibliography.bib}

\end{document}

%% file: figure1_2.pdf_tex
\begingroup%
  \makeatletter%
  \providecommand\color[2][]{%
    \errmessage{(Inkscape) Color is used for the text in Inkscape, but the package 'color.sty' is not loaded}%
    \renewcommand\color[2][]{}%
  }%
  \providecommand\transparent[1]{%
    \errmessage{(Inkscape) Transparency is used (non-zero) for the text in Inkscape, but the package 'transparent.sty' is not loaded}%
    \renewcommand\transparent[1]{}%
  }%
  \providecommand\rotatebox[2]{#2}%
  \newcommand*\fsize{\dimexpr\f@size pt\relax}%
  \newcommand*\lineheight[1]{\fontsize{\fsize}{#1\fsize}\selectfont}%
  \ifx\svgwidth\undefined%
    \setlength{\unitlength}{525.73040265bp}%
    \ifx\svgscale\undefined%
      \relax%
    \else%
      \setlength{\unitlength}{\unitlength * \real{\svgscale}}%
    \fi%
  \else%
    \setlength{\unitlength}{\svgwidth}%
  \fi%
  \global\let\svgwidth\undefined%
  \global\let\svgscale\undefined%
  \makeatother%
  \begin{picture}(1,0.43019721)%
    \lineheight{1}%
    \setlength\tabcolsep{0pt}%
    \put(0,0){\includegraphics[width=\unitlength,page=1]{figure1_2.pdf}}%
    \put(0.82553615,0.18041316){\color[rgb]{0,0,0}\makebox(0,0)[lt]{\lineheight{1.25}\smash{\begin{tabular}[t]{l}2\end{tabular}}}}%
    \put(0.72022899,0.28590879){\color[rgb]{0,0,0}\makebox(0,0)[lt]{\lineheight{1.25}\smash{\begin{tabular}[t]{l}3\end{tabular}}}}%
    \put(0.7729133,0.23322454){\color[rgb]{0,0,0}\makebox(0,0)[lt]{\lineheight{1.25}\smash{\begin{tabular}[t]{l}3\end{tabular}}}}%
    \put(0.72027407,0.23322861){\color[rgb]{0,0,0}\makebox(0,0)[lt]{\lineheight{1.25}\smash{\begin{tabular}[t]{l}0\end{tabular}}}}%
    \put(0.93077355,0.28588831){\color[rgb]{0,0,0}\makebox(0,0)[lt]{\lineheight{1.25}\smash{\begin{tabular}[t]{l}1\end{tabular}}}}%
    \put(0.87808932,0.1278356){\color[rgb]{0,0,0}\makebox(0,0)[lt]{\lineheight{1.25}\smash{\begin{tabular}[t]{l}1\end{tabular}}}}%
    \put(0.72027416,0.07517584){\color[rgb]{0,0,0}\makebox(0,0)[lt]{\lineheight{1.25}\smash{\begin{tabular}[t]{l}0\end{tabular}}}}%
    \put(0.77295847,0.07517584){\color[rgb]{0,0,0}\makebox(0,0)[lt]{\lineheight{1.25}\smash{\begin{tabular}[t]{l}0\end{tabular}}}}%
    \put(0.8256427,0.07517576){\color[rgb]{0,0,0}\makebox(0,0)[lt]{\lineheight{1.25}\smash{\begin{tabular}[t]{l}0\end{tabular}}}}%
    \put(0.87832709,0.07517593){\color[rgb]{0,0,0}\makebox(0,0)[lt]{\lineheight{1.25}\smash{\begin{tabular}[t]{l}0\end{tabular}}}}%
    \put(0.93101132,0.07517589){\color[rgb]{0,0,0}\makebox(0,0)[lt]{\lineheight{1.25}\smash{\begin{tabular}[t]{l}0\end{tabular}}}}%
    \put(0.72027416,0.12786007){\color[rgb]{0,0,0}\makebox(0,0)[lt]{\lineheight{1.25}\smash{\begin{tabular}[t]{l}0\end{tabular}}}}%
    \put(0.77295839,0.12786007){\color[rgb]{0,0,0}\makebox(0,0)[lt]{\lineheight{1.25}\smash{\begin{tabular}[t]{l}0\end{tabular}}}}%
    \put(0.8256427,0.12786007){\color[rgb]{0,0,0}\makebox(0,0)[lt]{\lineheight{1.25}\smash{\begin{tabular}[t]{l}0\end{tabular}}}}%
    \put(0.93101132,0.12786016){\color[rgb]{0,0,0}\makebox(0,0)[lt]{\lineheight{1.25}\smash{\begin{tabular}[t]{l}0\end{tabular}}}}%
    \put(0.93077355,0.18051979){\color[rgb]{0,0,0}\makebox(0,0)[lt]{\lineheight{1.25}\smash{\begin{tabular}[t]{l}1\end{tabular}}}}%
    \put(0.8256427,0.23322861){\color[rgb]{0,0,0}\makebox(0,0)[lt]{\lineheight{1.25}\smash{\begin{tabular}[t]{l}0\end{tabular}}}}%
    \put(0,0){\includegraphics[width=\unitlength,page=2]{figure1_2.pdf}}%
    \put(0.82540501,0.28588831){\color[rgb]{0,0,0}\makebox(0,0)[lt]{\lineheight{1.25}\smash{\begin{tabular}[t]{l}1\end{tabular}}}}%
    \put(0.77295839,0.28591286){\color[rgb]{0,0,0}\makebox(0,0)[lt]{\lineheight{1.25}\smash{\begin{tabular}[t]{l}0\end{tabular}}}}%
    \put(0.87832701,0.28591291){\color[rgb]{0,0,0}\makebox(0,0)[lt]{\lineheight{1.25}\smash{\begin{tabular}[t]{l}0\end{tabular}}}}%
    \put(0.77295847,0.18054434){\color[rgb]{0,0,0}\makebox(0,0)[lt]{\lineheight{1.25}\smash{\begin{tabular}[t]{l}0\end{tabular}}}}%
    \put(0.72027407,0.18054434){\color[rgb]{0,0,0}\makebox(0,0)[lt]{\lineheight{1.25}\smash{\begin{tabular}[t]{l}0\end{tabular}}}}%
    \put(0.39030106,0.17055985){\color[rgb]{0,0,0}\makebox(0,0)[lt]{\lineheight{1.25}\smash{\begin{tabular}[t]{l}\Large$\star$\end{tabular}}}}%
    \put(0.62305078,0.17117636){\color[rgb]{0,0,0}\makebox(0,0)[lt]{\lineheight{1.25}\smash{\begin{tabular}[t]{l}=\end{tabular}}}}%
    \put(0.1778793,0.38628523){\color[rgb]{0,0,0}\makebox(0,0)[lt]{\lineheight{1.25}\smash{\begin{tabular}[t]{l}\large$f$\end{tabular}}}}%
    \put(0.51114024,0.28479377){\color[rgb]{0,0,0}\makebox(0,0)[lt]{\lineheight{1.25}\smash{\begin{tabular}[t]{l}\large$\psi$\end{tabular}}}}%
    \put(0,0){\includegraphics[width=\unitlength,page=3]{figure1_2.pdf}}%
    \put(0.62305078,0.17117636){\color[rgb]{0,0,0}\makebox(0,0)[lt]{\lineheight{1.25}\smash{\begin{tabular}[t]{l}=\end{tabular}}}}%
    \put(0,0){\includegraphics[width=\unitlength,page=4]{figure1_2.pdf}}%
    \put(0.7641004,0.34804497){\makebox(0,0)[lt]{\lineheight{1.25}\smash{\begin{tabular}[t]{l}\large$f\star\psi$\end{tabular}}}}%
  \end{picture}%
\endgroup%

%% file: figure10.pdf_tex
\begingroup%
  \makeatletter%
  \providecommand\color[2][]{%
    \errmessage{(Inkscape) Color is used for the text in Inkscape, but the package 'color.sty' is not loaded}%
    \renewcommand\color[2][]{}%
  }%
  \providecommand\transparent[1]{%
    \errmessage{(Inkscape) Transparency is used (non-zero) for the text in Inkscape, but the package 'transparent.sty' is not loaded}%
    \renewcommand\transparent[1]{}%
  }%
  \providecommand\rotatebox[2]{#2}%
  \newcommand*\fsize{\dimexpr\f@size pt\relax}%
  \newcommand*\lineheight[1]{\fontsize{\fsize}{#1\fsize}\selectfont}%
  \ifx\svgwidth\undefined%
    \setlength{\unitlength}{550.38884414bp}%
    \ifx\svgscale\undefined%
      \relax%
    \else%
      \setlength{\unitlength}{\unitlength * \real{\svgscale}}%
    \fi%
  \else%
    \setlength{\unitlength}{\svgwidth}%
  \fi%
  \global\let\svgwidth\undefined%
  \global\let\svgscale\undefined%
  \makeatother%
  \begin{picture}(1,0.96106525)%
    \lineheight{1}%
    \setlength\tabcolsep{0pt}%
    \put(0,0){\includegraphics[width=\unitlength,page=1]{figure10.pdf}}%
    \put(0.46265615,0.70720598){\makebox(0,0)[lt]{\lineheight{1.25}\smash{\begin{tabular}[t]{l}$\star\psi$\end{tabular}}}}%
    \put(0,0){\includegraphics[width=\unitlength,page=2]{figure10.pdf}}%
    \put(0.46265615,0.28874766){\makebox(0,0)[lt]{\lineheight{1.25}\smash{\begin{tabular}[t]{l}$\star\psi$\end{tabular}}}}%
    \put(0,0){\includegraphics[width=\unitlength,page=3]{figure10.pdf}}%
    \put(0.79173626,0.49303091){\makebox(0,0)[lt]{\lineheight{1.25}\smash{\begin{tabular}[t]{l}$T_t$\end{tabular}}}}%
    \put(0,0){\includegraphics[width=\unitlength,page=4]{figure10.pdf}}%
    \put(0.12216668,0.49151187){\makebox(0,0)[lt]{\lineheight{1.25}\smash{\begin{tabular}[t]{l}$T_t$\end{tabular}}}}%
    \put(0.17309617,0.94149098){\makebox(0,0)[lt]{\lineheight{1.25}\smash{\begin{tabular}[t]{l}$f$\end{tabular}}}}%
    \put(0.17258044,0.00511713){\makebox(0,0)[lt]{\lineheight{1.25}\smash{\begin{tabular}[t]{l}$T_t(f)$\end{tabular}}}}%
    \put(0.78537852,0.11085275){\makebox(0,0)[lt]{\lineheight{1.25}\smash{\begin{tabular}[t]{l}$T_t(f\star\psi)$\end{tabular}}}}%
    \put(0.78585381,0.84768835){\makebox(0,0)[lt]{\lineheight{1.25}\smash{\begin{tabular}[t]{l}$f\star\psi$\end{tabular}}}}%
    \put(0,0){\includegraphics[width=\unitlength,page=5]{figure10.pdf}}%
  \end{picture}%
\endgroup%

%% file: figure11.pdf_tex
\begingroup%
  \makeatletter%
  \providecommand\color[2][]{%
    \errmessage{(Inkscape) Color is used for the text in Inkscape, but the package 'color.sty' is not loaded}%
    \renewcommand\color[2][]{}%
  }%
  \providecommand\transparent[1]{%
    \errmessage{(Inkscape) Transparency is used (non-zero) for the text in Inkscape, but the package 'transparent.sty' is not loaded}%
    \renewcommand\transparent[1]{}%
  }%
  \providecommand\rotatebox[2]{#2}%
  \newcommand*\fsize{\dimexpr\f@size pt\relax}%
  \newcommand*\lineheight[1]{\fontsize{\fsize}{#1\fsize}\selectfont}%
  \ifx\svgwidth\undefined%
    \setlength{\unitlength}{451.27675452bp}%
    \ifx\svgscale\undefined%
      \relax%
    \else%
      \setlength{\unitlength}{\unitlength * \real{\svgscale}}%
    \fi%
  \else%
    \setlength{\unitlength}{\svgwidth}%
  \fi%
  \global\let\svgwidth\undefined%
  \global\let\svgscale\undefined%
  \makeatother%
  \begin{picture}(1,0.40351166)%
    \lineheight{1}%
    \setlength\tabcolsep{0pt}%
    \put(0,0){\includegraphics[width=\unitlength,page=1]{figure11.pdf}}%
    \put(0.18943115,0.38338833){\makebox(0,0)[lt]{\lineheight{1.25}\smash{\begin{tabular}[t]{l}$e$\end{tabular}}}}%
    \put(0.36304404,0.19740902){\makebox(0,0)[lt]{\lineheight{1.25}\smash{\begin{tabular}[t]{l}$\,\,r^3$\end{tabular}}}}%
    \put(0.18803917,0.00460947){\makebox(0,0)[lt]{\lineheight{1.25}\smash{\begin{tabular}[t]{l}$r^2$\end{tabular}}}}%
    \put(-0.00197914,0.19739311){\makebox(0,0)[lt]{\lineheight{1.25}\smash{\begin{tabular}[t]{l}$r$\end{tabular}}}}%
    \put(0,0){\includegraphics[width=\unitlength,page=2]{figure11.pdf}}%
    \put(0.70288913,0.38338833){\makebox(0,0)[lt]{\lineheight{1.25}\smash{\begin{tabular}[t]{l}$e$\end{tabular}}}}%
    \put(0.87650197,0.19740902){\makebox(0,0)[lt]{\lineheight{1.25}\smash{\begin{tabular}[t]{l}$\,\,r^3$\end{tabular}}}}%
    \put(0.70149715,0.00460947){\makebox(0,0)[lt]{\lineheight{1.25}\smash{\begin{tabular}[t]{l}$r^2$\end{tabular}}}}%
    \put(0.51147875,0.19739306){\makebox(0,0)[lt]{\lineheight{1.25}\smash{\begin{tabular}[t]{l}$r$\end{tabular}}}}%
  \end{picture}%
\endgroup%

%% file: cosetpooling.pdf_tex
\begingroup%
  \makeatletter%
  \providecommand\color[2][]{%
    \errmessage{(Inkscape) Color is used for the text in Inkscape, but the package 'color.sty' is not loaded}%
    \renewcommand\color[2][]{}%
  }%
  \providecommand\transparent[1]{%
    \errmessage{(Inkscape) Transparency is used (non-zero) for the text in Inkscape, but the package 'transparent.sty' is not loaded}%
    \renewcommand\transparent[1]{}%
  }%
  \providecommand\rotatebox[2]{#2}%
  \newcommand*\fsize{\dimexpr\f@size pt\relax}%
  \newcommand*\lineheight[1]{\fontsize{\fsize}{#1\fsize}\selectfont}%
  \ifx\svgwidth\undefined%
    \setlength{\unitlength}{441.99196672bp}%
    \ifx\svgscale\undefined%
      \relax%
    \else%
      \setlength{\unitlength}{\unitlength * \real{\svgscale}}%
    \fi%
  \else%
    \setlength{\unitlength}{\svgwidth}%
  \fi%
  \global\let\svgwidth\undefined%
  \global\let\svgscale\undefined%
  \makeatother%
  \begin{picture}(1,0.53492111)%
    \lineheight{1}%
    \setlength\tabcolsep{0pt}%
    \put(0,0){\includegraphics[width=\unitlength,page=1]{cosetpooling.pdf}}%
    \put(0.24302733,0.46291352){\makebox(0,0)[lt]{\lineheight{1.25}\smash{\begin{tabular}[t]{l}$e$\end{tabular}}}}%
    \put(0.46350076,0.23764225){\makebox(0,0)[lt]{\lineheight{1.25}\smash{\begin{tabular}[t]{l}$  r^3$\end{tabular}}}}%
    \put(0.2430273,0.00558331){\makebox(0,0)[lt]{\lineheight{1.25}\smash{\begin{tabular}[t]{l}$r^2$\end{tabular}}}}%
    \put(-0.00239727,0.23762292){\makebox(0,0)[lt]{\lineheight{1.25}\smash{\begin{tabular}[t]{l}$r$\end{tabular}}}}%
    \put(0,0){\includegraphics[width=\unitlength,page=2]{cosetpooling.pdf}}%
    \put(0.5995365,0.25131537){\makebox(0,0)[lt]{\lineheight{1.25}\smash{\begin{tabular}[t]{l}$\underset{x\in(-2,2)C_4}{\max}f(x)$\end{tabular}}}}%
    \put(0,0){\includegraphics[width=\unitlength,page=3]{cosetpooling.pdf}}%
    \put(0.16548948,0.52099265){\makebox(0,0)[lt]{\lineheight{1.25}\smash{\begin{tabular}[t]{l}$(-2,2)$\end{tabular}}}}%
    \put(0,0){\includegraphics[width=\unitlength,page=4]{cosetpooling.pdf}}%
    \put(0.33055351,0.37816444){\makebox(0,0)[lt]{\lineheight{1.25}\smash{\begin{tabular}[t]{l}$(-2,2)r^3$\end{tabular}}}}%
    \put(0,0){\includegraphics[width=\unitlength,page=5]{cosetpooling.pdf}}%
    \put(0.17638684,0.23203816){\makebox(0,0)[lt]{\lineheight{1.25}\smash{\begin{tabular}[t]{l}$(-2,2)r^2$\end{tabular}}}}%
    \put(0,0){\includegraphics[width=\unitlength,page=6]{cosetpooling.pdf}}%
    \put(0.02854652,0.37773592){\makebox(0,0)[lt]{\lineheight{1.25}\smash{\begin{tabular}[t]{l}$(-2,2)r$\end{tabular}}}}%
    \put(0,0){\includegraphics[width=\unitlength,page=7]{cosetpooling.pdf}}%
    \put(0.85547642,0.37449595){\makebox(0,0)[lt]{\lineheight{1.25}\smash{\begin{tabular}[t]{l}$(-2,2)C_4$\end{tabular}}}}%
  \end{picture}%
\endgroup%

%% file: figure8.pdf_tex
\begingroup%
  \makeatletter%
  \providecommand\color[2][]{%
    \errmessage{(Inkscape) Color is used for the text in Inkscape, but the package 'color.sty' is not loaded}%
    \renewcommand\color[2][]{}%
  }%
  \providecommand\transparent[1]{%
    \errmessage{(Inkscape) Transparency is used (non-zero) for the text in Inkscape, but the package 'transparent.sty' is not loaded}%
    \renewcommand\transparent[1]{}%
  }%
  \providecommand\rotatebox[2]{#2}%
  \newcommand*\fsize{\dimexpr\f@size pt\relax}%
  \newcommand*\lineheight[1]{\fontsize{\fsize}{#1\fsize}\selectfont}%
  \ifx\svgwidth\undefined%
    \setlength{\unitlength}{243.36127995bp}%
    \ifx\svgscale\undefined%
      \relax%
    \else%
      \setlength{\unitlength}{\unitlength * \real{\svgscale}}%
    \fi%
  \else%
    \setlength{\unitlength}{\svgwidth}%
  \fi%
  \global\let\svgwidth\undefined%
  \global\let\svgscale\undefined%
  \makeatother%
  \begin{picture}(1,0.52857856)%
    \lineheight{1}%
    \setlength\tabcolsep{0pt}%
    \put(0,0){\includegraphics[width=\unitlength,page=1]{figure8.pdf}}%
  \end{picture}%
\endgroup%

%% file: figure7_2.pdf_tex
\begingroup%
  \makeatletter%
  \providecommand\color[2][]{%
    \errmessage{(Inkscape) Color is used for the text in Inkscape, but the package 'color.sty' is not loaded}%
    \renewcommand\color[2][]{}%
  }%
  \providecommand\transparent[1]{%
    \errmessage{(Inkscape) Transparency is used (non-zero) for the text in Inkscape, but the package 'transparent.sty' is not loaded}%
    \renewcommand\transparent[1]{}%
  }%
  \providecommand\rotatebox[2]{#2}%
  \newcommand*\fsize{\dimexpr\f@size pt\relax}%
  \newcommand*\lineheight[1]{\fontsize{\fsize}{#1\fsize}\selectfont}%
  \ifx\svgwidth\undefined%
    \setlength{\unitlength}{193.46746742bp}%
    \ifx\svgscale\undefined%
      \relax%
    \else%
      \setlength{\unitlength}{\unitlength * \real{\svgscale}}%
    \fi%
  \else%
    \setlength{\unitlength}{\svgwidth}%
  \fi%
  \global\let\svgwidth\undefined%
  \global\let\svgscale\undefined%
  \makeatother%
  \begin{picture}(1,0.85518395)%
    \lineheight{1}%
    \setlength\tabcolsep{0pt}%
    \put(0,0){\includegraphics[width=\unitlength,page=1]{figure7_2.pdf}}%
    \put(0.40811146,0.70040753){\makebox(0,0)[lt]{\lineheight{1.25}\smash{\begin{tabular}[t]{l}$\pi_0(r)$\end{tabular}}}}%
    \put(0,0){\includegraphics[width=\unitlength,page=2]{figure7_2.pdf}}%
    \put(0.40918121,0.10230051){\makebox(0,0)[lt]{\lineheight{1.25}\smash{\begin{tabular}[t]{l}$\rho(r)$\end{tabular}}}}%
    \put(0,0){\includegraphics[width=\unitlength,page=3]{figure7_2.pdf}}%
    \put(0.09192066,0.35624035){\makebox(0,0)[lt]{\lineheight{1.25}\smash{\begin{tabular}[t]{l}$\Psi$\end{tabular}}}}%
    \put(0.73017853,0.35536843){\makebox(0,0)[lt]{\lineheight{1.25}\smash{\begin{tabular}[t]{l}$\Psi$\end{tabular}}}}%
  \end{picture}%
\endgroup%

%% file: figure9_2.pdf_tex
\begingroup%
  \makeatletter%
  \providecommand\color[2][]{%
    \errmessage{(Inkscape) Color is used for the text in Inkscape, but the package 'color.sty' is not loaded}%
    \renewcommand\color[2][]{}%
  }%
  \providecommand\transparent[1]{%
    \errmessage{(Inkscape) Transparency is used (non-zero) for the text in Inkscape, but the package 'transparent.sty' is not loaded}%
    \renewcommand\transparent[1]{}%
  }%
  \providecommand\rotatebox[2]{#2}%
  \newcommand*\fsize{\dimexpr\f@size pt\relax}%
  \newcommand*\lineheight[1]{\fontsize{\fsize}{#1\fsize}\selectfont}%
  \ifx\svgwidth\undefined%
    \setlength{\unitlength}{199.04140135bp}%
    \ifx\svgscale\undefined%
      \relax%
    \else%
      \setlength{\unitlength}{\unitlength * \real{\svgscale}}%
    \fi%
  \else%
    \setlength{\unitlength}{\svgwidth}%
  \fi%
  \global\let\svgwidth\undefined%
  \global\let\svgscale\undefined%
  \makeatother%
  \begin{picture}(1,0.56966199)%
    \lineheight{1}%
    \setlength\tabcolsep{0pt}%
    \put(0,0){\includegraphics[width=\unitlength,page=1]{figure9_2.pdf}}%
    \put(0.27113439,0.43783918){\makebox(0,0)[lt]{\lineheight{1.25}\smash{\begin{tabular}[t]{l}$\pi_0(r)$\end{tabular}}}}%
    \put(0,0){\includegraphics[width=\unitlength,page=2]{figure9_2.pdf}}%
    \put(0.27113439,0.12110708){\makebox(0,0)[lt]{\lineheight{1.25}\smash{\begin{tabular}[t]{l}$\pi_0(t)$\end{tabular}}}}%
    \put(0,0){\includegraphics[width=\unitlength,page=3]{figure9_2.pdf}}%
  \end{picture}%
\endgroup%

%% file: figure2.pdf_tex
\begingroup%
  \makeatletter%
  \providecommand\color[2][]{%
    \errmessage{(Inkscape) Color is used for the text in Inkscape, but the package 'color.sty' is not loaded}%
    \renewcommand\color[2][]{}%
  }%
  \providecommand\transparent[1]{%
    \errmessage{(Inkscape) Transparency is used (non-zero) for the text in Inkscape, but the package 'transparent.sty' is not loaded}%
    \renewcommand\transparent[1]{}%
  }%
  \providecommand\rotatebox[2]{#2}%
  \newcommand*\fsize{\dimexpr\f@size pt\relax}%
  \newcommand*\lineheight[1]{\fontsize{\fsize}{#1\fsize}\selectfont}%
  \ifx\svgwidth\undefined%
    \setlength{\unitlength}{627.8057332bp}%
    \ifx\svgscale\undefined%
      \relax%
    \else%
      \setlength{\unitlength}{\unitlength * \real{\svgscale}}%
    \fi%
  \else%
    \setlength{\unitlength}{\svgwidth}%
  \fi%
  \global\let\svgwidth\undefined%
  \global\let\svgscale\undefined%
  \makeatother%
  \begin{picture}(1,0.07256814)%
    \lineheight{1}%
    \setlength\tabcolsep{0pt}%
    \put(0,0){\includegraphics[width=\unitlength,page=1]{figure2.pdf}}%
    \put(0.22300146,0.00410465){\color[rgb]{0,0,0}\transparent{0}\makebox(0,0)[lt]{\lineheight{1.25}\smash{\begin{tabular}[t]{l}                \end{tabular}}}}%
    \put(0.22363911,0.0040288){\makebox(0,0)[lt]{\lineheight{1.25}\smash{\begin{tabular}[t]{l},\end{tabular}}}}%
    \put(0.30488224,0.0040288){\makebox(0,0)[lt]{\lineheight{1.25}\smash{\begin{tabular}[t]{l},\end{tabular}}}}%
    \put(0.38322503,0.0040288){\makebox(0,0)[lt]{\lineheight{1.25}\smash{\begin{tabular}[t]{l},\end{tabular}}}}%
    \put(0.46362586,0.0040288){\makebox(0,0)[lt]{\lineheight{1.25}\smash{\begin{tabular}[t]{l},\end{tabular}}}}%
    \put(0.54312143,0.0040288){\makebox(0,0)[lt]{\lineheight{1.25}\smash{\begin{tabular}[t]{l},\end{tabular}}}}%
    \put(0.62638839,0.0040288){\makebox(0,0)[lt]{\lineheight{1.25}\smash{\begin{tabular}[t]{l},\end{tabular}}}}%
    \put(0.70814671,0.0040288){\makebox(0,0)[lt]{\lineheight{1.25}\smash{\begin{tabular}[t]{l},\end{tabular}}}}%
    \put(0.79005598,0.0040288){\makebox(0,0)[lt]{\lineheight{1.25}\smash{\begin{tabular}[t]{l},\end{tabular}}}}%
    \put(-0.00076221,0.02257476){\makebox(0,0)[lt]{\lineheight{1.25}\smash{\begin{tabular}[t]{l}\Huge$\mathcal{C}=\{$\end{tabular}}}}%
    \put(0.88066047,0.02243793){\makebox(0,0)[lt]{\lineheight{1.25}\smash{\begin{tabular}[t]{l}\Huge$\}$\end{tabular}}}}%
  \end{picture}%
\endgroup%

%% file: figure3.pdf_tex
\begingroup%
  \makeatletter%
  \providecommand\color[2][]{%
    \errmessage{(Inkscape) Color is used for the text in Inkscape, but the package 'color.sty' is not loaded}%
    \renewcommand\color[2][]{}%
  }%
  \providecommand\transparent[1]{%
    \errmessage{(Inkscape) Transparency is used (non-zero) for the text in Inkscape, but the package 'transparent.sty' is not loaded}%
    \renewcommand\transparent[1]{}%
  }%
  \providecommand\rotatebox[2]{#2}%
  \newcommand*\fsize{\dimexpr\f@size pt\relax}%
  \newcommand*\lineheight[1]{\fontsize{\fsize}{#1\fsize}\selectfont}%
  \ifx\svgwidth\undefined%
    \setlength{\unitlength}{564.49349282bp}%
    \ifx\svgscale\undefined%
      \relax%
    \else%
      \setlength{\unitlength}{\unitlength * \real{\svgscale}}%
    \fi%
  \else%
    \setlength{\unitlength}{\svgwidth}%
  \fi%
  \global\let\svgwidth\undefined%
  \global\let\svgscale\undefined%
  \makeatother%
  \begin{picture}(1,0.07673441)%
    \lineheight{1}%
    \setlength\tabcolsep{0pt}%
    \put(0,0){\includegraphics[width=\unitlength,page=1]{figure3.pdf}}%
    \put(0.46774163,0.04338576){\makebox(0,0)[lt]{\lineheight{1.25}\smash{\begin{tabular}[t]{l}$\pi_0(r)$\end{tabular}}}}%
    \put(0,0){\includegraphics[width=\unitlength,page=2]{figure3.pdf}}%
  \end{picture}%
\endgroup%

%% file: filtertrans.pdf_tex
\begingroup%
  \makeatletter%
  \providecommand\color[2][]{%
    \errmessage{(Inkscape) Color is used for the text in Inkscape, but the package 'color.sty' is not loaded}%
    \renewcommand\color[2][]{}%
  }%
  \providecommand\transparent[1]{%
    \errmessage{(Inkscape) Transparency is used (non-zero) for the text in Inkscape, but the package 'transparent.sty' is not loaded}%
    \renewcommand\transparent[1]{}%
  }%
  \providecommand\rotatebox[2]{#2}%
  \newcommand*\fsize{\dimexpr\f@size pt\relax}%
  \newcommand*\lineheight[1]{\fontsize{\fsize}{#1\fsize}\selectfont}%
  \ifx\svgwidth\undefined%
    \setlength{\unitlength}{450.02059011bp}%
    \ifx\svgscale\undefined%
      \relax%
    \else%
      \setlength{\unitlength}{\unitlength * \real{\svgscale}}%
    \fi%
  \else%
    \setlength{\unitlength}{\svgwidth}%
  \fi%
  \global\let\svgwidth\undefined%
  \global\let\svgscale\undefined%
  \makeatother%
  \begin{picture}(1,0.5992084)%
    \lineheight{1}%
    \setlength\tabcolsep{0pt}%
    \put(0,0){\includegraphics[width=\unitlength,page=1]{filtertrans.pdf}}%
    \put(0.09551078,0.4944408){\color[rgb]{0,0,0}\transparent{0}\makebox(0,0)[lt]{\lineheight{1.25}\smash{\begin{tabular}[t]{l}                \end{tabular}}}}%
    \put(0,0){\includegraphics[width=\unitlength,page=2]{filtertrans.pdf}}%
    \put(0.09551078,0.01113181){\color[rgb]{0,0,0}\transparent{0}\makebox(0,0)[lt]{\lineheight{1.25}\smash{\begin{tabular}[t]{l}                \end{tabular}}}}%
    \put(0,0){\includegraphics[width=\unitlength,page=3]{filtertrans.pdf}}%
    \put(0.67217827,0.37438375){\makebox(0,0)[lt]{\lineheight{1.25}\smash{\begin{tabular}[t]{l} \end{tabular}}}}%
    \put(0,0){\includegraphics[width=\unitlength,page=4]{filtertrans.pdf}}%
  \end{picture}%
\endgroup%

%% file: figure6.pdf_tex
\begingroup%
  \makeatletter%
  \providecommand\color[2][]{%
    \errmessage{(Inkscape) Color is used for the text in Inkscape, but the package 'color.sty' is not loaded}%
    \renewcommand\color[2][]{}%
  }%
  \providecommand\transparent[1]{%
    \errmessage{(Inkscape) Transparency is used (non-zero) for the text in Inkscape, but the package 'transparent.sty' is not loaded}%
    \renewcommand\transparent[1]{}%
  }%
  \providecommand\rotatebox[2]{#2}%
  \newcommand*\fsize{\dimexpr\f@size pt\relax}%
  \newcommand*\lineheight[1]{\fontsize{\fsize}{#1\fsize}\selectfont}%
  \ifx\svgwidth\undefined%
    \setlength{\unitlength}{646.91710681bp}%
    \ifx\svgscale\undefined%
      \relax%
    \else%
      \setlength{\unitlength}{\unitlength * \real{\svgscale}}%
    \fi%
  \else%
    \setlength{\unitlength}{\svgwidth}%
  \fi%
  \global\let\svgwidth\undefined%
  \global\let\svgscale\undefined%
  \makeatother%
  \begin{picture}(1,0.06695765)%
    \lineheight{1}%
    \setlength\tabcolsep{0pt}%
    \put(0,0){\includegraphics[width=\unitlength,page=1]{figure6.pdf}}%
    \put(0.17020554,0.03726173){\makebox(0,0)[lt]{\lineheight{1.25}\smash{\begin{tabular}[t]{l}$\pi_0(r)$\end{tabular}}}}%
    \put(0,0){\includegraphics[width=\unitlength,page=2]{figure6.pdf}}%
    \put(0.65877208,0.03726178){\makebox(0,0)[lt]{\lineheight{1.25}\smash{\begin{tabular}[t]{l}$\pi_0(r)$\end{tabular}}}}%
    \put(0,0){\includegraphics[width=\unitlength,page=3]{figure6.pdf}}%
    \put(-0.0012869,0.02796065){\makebox(0,0)[lt]{\lineheight{1.25}\smash{\begin{tabular}[t]{l}$\psi_1=$\end{tabular}}}}%
    \put(0.32798726,0.02796065){\makebox(0,0)[lt]{\lineheight{1.25}\smash{\begin{tabular}[t]{l}$=[1]\cdot\psi_1$,\end{tabular}}}}%
    \put(0.48514736,0.02796065){\makebox(0,0)[lt]{\lineheight{1.25}\smash{\begin{tabular}[t]{l}$\psi_4=$\end{tabular}}}}%
    \put(0.83381498,0.02796065){\makebox(0,0)[lt]{\lineheight{1.25}\smash{\begin{tabular}[t]{l}$=[-1]\cdot\psi_4$\end{tabular}}}}%
  \end{picture}%
\endgroup%

%% file: figure5.pdf_tex
\begingroup%
  \makeatletter%
  \providecommand\color[2][]{%
    \errmessage{(Inkscape) Color is used for the text in Inkscape, but the package 'color.sty' is not loaded}%
    \renewcommand\color[2][]{}%
  }%
  \providecommand\transparent[1]{%
    \errmessage{(Inkscape) Transparency is used (non-zero) for the text in Inkscape, but the package 'transparent.sty' is not loaded}%
    \renewcommand\transparent[1]{}%
  }%
  \providecommand\rotatebox[2]{#2}%
  \newcommand*\fsize{\dimexpr\f@size pt\relax}%
  \newcommand*\lineheight[1]{\fontsize{\fsize}{#1\fsize}\selectfont}%
  \ifx\svgwidth\undefined%
    \setlength{\unitlength}{416.41571525bp}%
    \ifx\svgscale\undefined%
      \relax%
    \else%
      \setlength{\unitlength}{\unitlength * \real{\svgscale}}%
    \fi%
  \else%
    \setlength{\unitlength}{\svgwidth}%
  \fi%
  \global\let\svgwidth\undefined%
  \global\let\svgscale\undefined%
  \makeatother%
  \begin{picture}(1,0.60269423)%
    \lineheight{1}%
    \setlength\tabcolsep{0pt}%
    \put(0,0){\includegraphics[width=\unitlength,page=1]{figure5.pdf}}%
    \put(0.72093217,0.49327013){\color[rgb]{0,0,0}\makebox(0,0)[lt]{\lineheight{1.25}\smash{\begin{tabular}[t]{l}\small A1\end{tabular}}}}%
    \put(0.72093217,0.4318733){\color[rgb]{0,0,0}\makebox(0,0)[lt]{\lineheight{1.25}\smash{\begin{tabular}[t]{l}A1\end{tabular}}}}%
    \put(0.72093217,0.37139281){\color[rgb]{0,0,0}\makebox(0,0)[lt]{\lineheight{1.25}\smash{\begin{tabular}[t]{l}A2\end{tabular}}}}%
    \put(0.71860998,0.31230313){\color[rgb]{0,0,0}\makebox(0,0)[lt]{\lineheight{1.25}\smash{\begin{tabular}[t]{l}B1\end{tabular}}}}%
    \put(0.71860998,0.25359349){\color[rgb]{0,0,0}\makebox(0,0)[lt]{\lineheight{1.25}\smash{\begin{tabular}[t]{l}B2\end{tabular}}}}%
    \put(0,0){\includegraphics[width=\unitlength,page=2]{figure5.pdf}}%
    \put(0.71860998,0.19505652){\color[rgb]{0,0,0}\makebox(0,0)[lt]{\lineheight{1.25}\smash{\begin{tabular}[t]{l}E\end{tabular}}}}%
    \put(0.71860998,0.13807917){\color[rgb]{0,0,0}\makebox(0,0)[lt]{\lineheight{1.25}\smash{\begin{tabular}[t]{l}E\end{tabular}}}}%
    \put(0,0){\includegraphics[width=\unitlength,page=3]{figure5.pdf}}%
    \put(-0.00218791,0.40668727){\makebox(0,0)[lt]{\lineheight{1.25}\smash{\begin{tabular}[t]{l}$m_1=3$\end{tabular}}}}%
    \put(-0.00218791,0.32138569){\makebox(0,0)[lt]{\lineheight{1.25}\smash{\begin{tabular}[t]{l}$m_2=0$\end{tabular}}}}%
    \put(-0.00263954,0.25251775){\makebox(0,0)[lt]{\lineheight{1.25}\smash{\begin{tabular}[t]{l}$m_3=1$\end{tabular}}}}%
    \put(-0.00258537,0.19360258){\makebox(0,0)[lt]{\lineheight{1.25}\smash{\begin{tabular}[t]{l}$m_4=1$\end{tabular}}}}%
    \put(0,0){\includegraphics[width=\unitlength,page=4]{figure5.pdf}}%
    \put(-0.00309761,0.10008859){\makebox(0,0)[lt]{\lineheight{1.25}\smash{\begin{tabular}[t]{l}$m_5=2$\end{tabular}}}}%
    \put(0.20570783,0.50167274){\makebox(0,0)[lt]{\lineheight{1.25}\smash{\begin{tabular}[t]{l}$(\mathcal{F}_0,\pi_0)$\end{tabular}}}}%
    \put(0.78813972,0.46362226){\makebox(0,0)[lt]{\lineheight{1.25}\smash{\begin{tabular}[t]{l}$m_1'=2$\end{tabular}}}}%
    \put(0.78824359,0.37267387){\makebox(0,0)[lt]{\lineheight{1.25}\smash{\begin{tabular}[t]{l}$m_2'=1$\end{tabular}}}}%
    \put(0.78824359,0.31453377){\makebox(0,0)[lt]{\lineheight{1.25}\smash{\begin{tabular}[t]{l}$m_3'=1$\end{tabular}}}}%
    \put(0.78824359,0.25215805){\makebox(0,0)[lt]{\lineheight{1.25}\smash{\begin{tabular}[t]{l}$m_4'=1$\end{tabular}}}}%
    \put(0.78824359,0.16533555){\makebox(0,0)[lt]{\lineheight{1.25}\smash{\begin{tabular}[t]{l}$m_5'=1$\end{tabular}}}}%
  \end{picture}%
\endgroup%

%% file: figure13.pdf_tex
\begingroup%
  \makeatletter%
  \providecommand\color[2][]{%
    \errmessage{(Inkscape) Color is used for the text in Inkscape, but the package 'color.sty' is not loaded}%
    \renewcommand\color[2][]{}%
  }%
  \providecommand\transparent[1]{%
    \errmessage{(Inkscape) Transparency is used (non-zero) for the text in Inkscape, but the package 'transparent.sty' is not loaded}%
    \renewcommand\transparent[1]{}%
  }%
  \providecommand\rotatebox[2]{#2}%
  \newcommand*\fsize{\dimexpr\f@size pt\relax}%
  \newcommand*\lineheight[1]{\fontsize{\fsize}{#1\fsize}\selectfont}%
  \ifx\svgwidth\undefined%
    \setlength{\unitlength}{252.25934135bp}%
    \ifx\svgscale\undefined%
      \relax%
    \else%
      \setlength{\unitlength}{\unitlength * \real{\svgscale}}%
    \fi%
  \else%
    \setlength{\unitlength}{\svgwidth}%
  \fi%
  \global\let\svgwidth\undefined%
  \global\let\svgscale\undefined%
  \makeatother%
  \begin{picture}(1,0.68354531)%
    \lineheight{1}%
    \setlength\tabcolsep{0pt}%
    \put(0,0){\includegraphics[width=\unitlength,page=1]{figure13.pdf}}%
    \put(0.15973211,0.22861853){\makebox(0,0)[lt]{\lineheight{1.25}\smash{\begin{tabular}[t]{l}$e$\end{tabular}}}}%
    \put(0.15973211,0.00857432){\makebox(0,0)[lt]{\lineheight{1.25}\smash{\begin{tabular}[t]{l}$r^2$\end{tabular}}}}%
    \put(0.05088368,0.1314519){\makebox(0,0)[lt]{\lineheight{1.25}\smash{\begin{tabular}[t]{l}$r$\end{tabular}}}}%
    \put(0.25374553,0.13145207){\makebox(0,0)[lt]{\lineheight{1.25}\smash{\begin{tabular}[t]{l}$r^3$\end{tabular}}}}%
    \put(0,0){\includegraphics[width=\unitlength,page=2]{figure13.pdf}}%
    \put(0.75588245,0.22861853){\makebox(0,0)[lt]{\lineheight{1.25}\smash{\begin{tabular}[t]{l}$e$\end{tabular}}}}%
    \put(0.75588245,0.00857398){\makebox(0,0)[lt]{\lineheight{1.25}\smash{\begin{tabular}[t]{l}$r^2$\end{tabular}}}}%
    \put(0.64703402,0.13145155){\makebox(0,0)[lt]{\lineheight{1.25}\smash{\begin{tabular}[t]{l}$r$\end{tabular}}}}%
    \put(0.84989604,0.13145172){\makebox(0,0)[lt]{\lineheight{1.25}\smash{\begin{tabular}[t]{l}$r^3$\end{tabular}}}}%
  \end{picture}%
\endgroup%

%% file: s3filters2.pdf_tex
\begingroup%
  \makeatletter%
  \providecommand\color[2][]{%
    \errmessage{(Inkscape) Color is used for the text in Inkscape, but the package 'color.sty' is not loaded}%
    \renewcommand\color[2][]{}%
  }%
  \providecommand\transparent[1]{%
    \errmessage{(Inkscape) Transparency is used (non-zero) for the text in Inkscape, but the package 'transparent.sty' is not loaded}%
    \renewcommand\transparent[1]{}%
  }%
  \providecommand\rotatebox[2]{#2}%
  \newcommand*\fsize{\dimexpr\f@size pt\relax}%
  \newcommand*\lineheight[1]{\fontsize{\fsize}{#1\fsize}\selectfont}%
  \ifx\svgwidth\undefined%
    \setlength{\unitlength}{445.18778963bp}%
    \ifx\svgscale\undefined%
      \relax%
    \else%
      \setlength{\unitlength}{\unitlength * \real{\svgscale}}%
    \fi%
  \else%
    \setlength{\unitlength}{\svgwidth}%
  \fi%
  \global\let\svgwidth\undefined%
  \global\let\svgscale\undefined%
  \makeatother%
  \begin{picture}(1,0.08676968)%
    \lineheight{1}%
    \setlength\tabcolsep{0pt}%
    \put(0,0){\includegraphics[width=\unitlength,page=1]{s3filters2.pdf}}%
  \end{picture}%
\endgroup%

%% file: s3filtersdecomp3.pdf_tex
\begingroup%
  \makeatletter%
  \providecommand\color[2][]{%
    \errmessage{(Inkscape) Color is used for the text in Inkscape, but the package 'color.sty' is not loaded}%
    \renewcommand\color[2][]{}%
  }%
  \providecommand\transparent[1]{%
    \errmessage{(Inkscape) Transparency is used (non-zero) for the text in Inkscape, but the package 'transparent.sty' is not loaded}%
    \renewcommand\transparent[1]{}%
  }%
  \providecommand\rotatebox[2]{#2}%
  \newcommand*\fsize{\dimexpr\f@size pt\relax}%
  \newcommand*\lineheight[1]{\fontsize{\fsize}{#1\fsize}\selectfont}%
  \ifx\svgwidth\undefined%
    \setlength{\unitlength}{426.40248961bp}%
    \ifx\svgscale\undefined%
      \relax%
    \else%
      \setlength{\unitlength}{\unitlength * \real{\svgscale}}%
    \fi%
  \else%
    \setlength{\unitlength}{\svgwidth}%
  \fi%
  \global\let\svgwidth\undefined%
  \global\let\svgscale\undefined%
  \makeatother%
  \begin{picture}(1,0.63206755)%
    \lineheight{1}%
    \setlength\tabcolsep{0pt}%
    \put(0,0){\includegraphics[width=\unitlength,page=1]{s3filtersdecomp3.pdf}}%
    \put(-0.00079178,0.52231625){\makebox(0,0)[lt]{\lineheight{1.25}\smash{\begin{tabular}[t]{l}id\end{tabular}}}}%
    \put(-0.00079177,0.30999187){\makebox(0,0)[lt]{\lineheight{1.25}\smash{\begin{tabular}[t]{l}sgn\end{tabular}}}}%
    \put(-0.00141995,0.09926783){\makebox(0,0)[lt]{\lineheight{1.25}\smash{\begin{tabular}[t]{l}$V_s$\end{tabular}}}}%
  \end{picture}%
\endgroup%